\documentclass[11pt,a4paper]{article}
\usepackage[hyperref]{acl2017}
\usepackage{mathptmx}
\usepackage{latexsym}
\usepackage{url}
\usepackage{amsmath}
\usepackage{pgfplots}
\usepackage{xcolor} 
\usepackage{boxedminipage} 
\usetikzlibrary{patterns} 

\definecolor{gred}{RGB}{219,68,55}
\definecolor{gblue}{RGB}{66,133,244}
\definecolor{gyellow}{RGB}{244,180,0}
\definecolor{ggreen}{RGB}{15,157,88}
\definecolor{ggrey}{RGB}{115,115,115}

\newcommand{\hlpgen}[2]{\fboxsep1pt \colorbox{ggreen!#2}{\strut #1}} 
\newcommand{\hlcov}[2]{\fboxsep1pt \colorbox{gyellow!#2}{\strut #1}} 
\newcommand{\error}[1]{\textcolor{gred}{\textbf{#1}}} 
\newcommand{\fph}[1]{\textcolor{gblue}{\textbf{#1}}} 
\newcommand{\reph}[1]{\textcolor{ggreen}{\textbf{#1}}} 
\newcommand{\novelh}[1]{\textcolor{gblue}{\textbf{#1}}} 

\pgfplotsset{compat=1.11,
        /pgfplots/ybar legend/.style={
        /pgfplots/legend image code/.code={
        \draw[##1,/tikz/.cd,bar width=3pt,yshift=-0.2em,bar shift=0pt]
                plot coordinates {(0cm,0.8em)};
                },
	},
}

\aclfinalcopy 
\def\aclpaperid{761} 

\newcommand{\pgen}{p_{\text{gen}}}
\newcommand{\pvocab}{P_{\text{vocab}}}

\title{Get To The Point: Summarization with Pointer-Generator Networks}

\author{Abigail See \\
Stanford University \\
  {\tt abisee@stanford.edu} \\\And
  Peter J. Liu \\
  Google Brain \\
  {\tt peterjliu@google.com} \\\And
	 Christopher D. Manning\\
 Stanford University \\
  {\tt manning@stanford.edu} \\}

\date{}

\begin{document}
\maketitle
\begin{abstract}
Neural sequence-to-sequence models have provided a viable new approach
for \textit{abstractive} text summarization (meaning they are not restricted to simply selecting and rearranging passages from the original text).
However, these models have two shortcomings: they are liable to reproduce factual details inaccurately, and they tend to repeat themselves.
In this work we propose a novel architecture that augments the standard sequence-to-sequence attentional model in two orthogonal ways.
First, we use a hybrid pointer-generator network that can copy words from the source text via \textit{pointing}, which aids accurate reproduction of information, while retaining the ability to produce novel words through the \textit{generator}.
Second, we use \textit{coverage} to keep track of what has been summarized, which discourages repetition.
We apply our model to the \textit{CNN / Daily Mail} summarization
task, outperforming the current abstractive state-of-the-art 
by at least 2 ROUGE points.
\end{abstract}


\begin{figure}[h]
\begin{boxedminipage}{\columnwidth}
\scriptsize
\textbf{Original Text (truncated):}
lagos, nigeria (cnn) a day after winning nigeria's presidency, \fph{\textit{muhammadu buhari}} told cnn's christiane amanpour that \fph{he plans to aggressively fight corruption that has long plagued nigeria} and go after the root of the nation's unrest. \textit{buhari} said he'll ``rapidly give attention'' to curbing violence in the northeast part of nigeria, where the terrorist group boko haram operates. by cooperating with neighboring nations chad, cameroon and niger, \fph{he said his administration is confident it will be able to thwart criminals} and others contributing to nigeria's instability. for the first time in nigeria's history, the opposition defeated the ruling party in democratic elections. \textit{buhari} defeated incumbent goodluck jonathan by about 2 million votes, according to nigeria's independent national electoral commission. \fph{the win comes after a long history of military rule, coups and botched attempts at democracy in africa's most populous nation.}
\newline \rule{\columnwidth}{1.2pt}
\textbf{Baseline Seq2Seq + Attention:}
\error{UNK UNK} says his administration is confident it will be able to \error{destabilize nigeria's economy}. \error{UNK} says his administration is confident it will be able to thwart criminals and other \error{nigerians}. \error{he says the country has long nigeria and nigeria's economy.}
\newline \rule{\columnwidth}{0.4pt}
\textbf{Pointer-Gen:}
\textit{muhammadu buhari} says he plans to aggressively fight corruption \reph{in the northeast part of nigeria}. he says he'll ``rapidly give attention'' to curbing violence \reph{in the northeast part of nigeria}.
he says his administration is confident it will be able to thwart criminals.
\newline \rule{\columnwidth}{0.4pt}
\textbf{Pointer-Gen + Coverage:}
\textit{muhammadu buhari} says he plans to aggressively fight corruption that has long plagued nigeria. he says his administration is confident it will be able to thwart criminals. the win comes after a long history of military rule, coups and botched attempts at democracy in africa's most populous nation.
\end{boxedminipage}
\caption{
Comparison of output of 3 abstractive summarization models on a news article.
The baseline model makes \error{factual errors}, a \error{nonsensical sentence} and struggles with OOV words \textit{muhammadu buhari}.
The pointer-generator model is accurate but \reph{repeats} \reph{itself}.
Coverage eliminates repetition.
The final summary is composed from \fph{several fragments}.
}
\label{fig_firstpage}
\end{figure}

\begin{figure*}[t]
\centering
\includegraphics[width=0.9\textwidth, trim={15 30 90 130},clip]{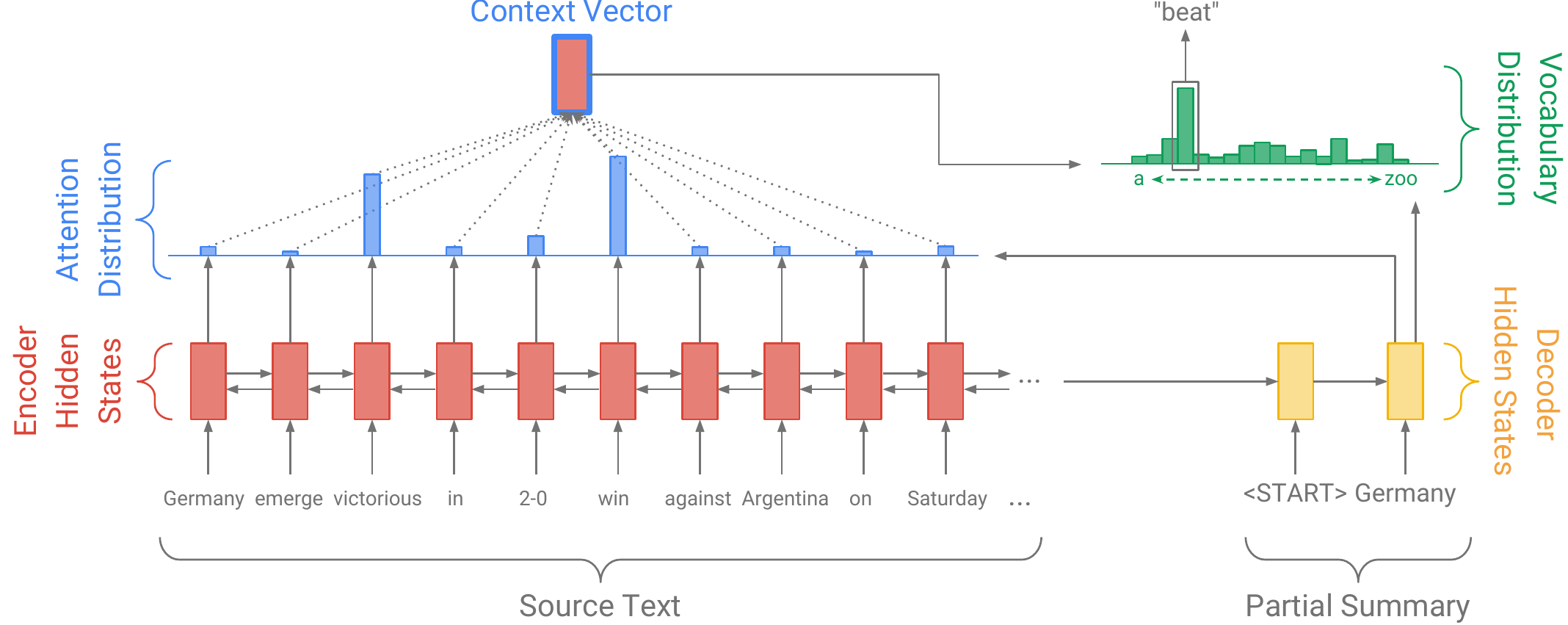} 
\caption{Baseline sequence-to-sequence model with attention. The model may attend to relevant words in the source text to generate novel words, e.g., to produce the novel word \textit{beat} in the abstractive summary \textit{Germany \textbf{beat} Argentina 2-0} the model may attend to the words \textit{victorious} and \textit{win} in the source text.}
\label{fig_vanilla_diagram}
\vspace*{-0.5ex}
\end{figure*}

\begin{section}{Introduction}
Summarization is the task of condensing a piece of text to a shorter version that contains the main information from the original.
There are two broad approaches to summarization: \emph{extractive} and \emph{abstractive}.
\textit{Extractive methods} assemble summaries exclusively from
passages (usually whole sentences) taken directly from the source
text, while \textit{abstractive methods} may generate novel words and
phrases not featured in the source text -- as a human-written
abstract usually does.
The extractive approach is easier, because copying large chunks of text from the source document ensures baseline levels of grammaticality and accuracy.
On the other hand, sophisticated abilities that are crucial to high-quality summarization, such as paraphrasing, generalization, or the incorporation of real-world knowledge, are possible only in an abstractive framework (see Figure \ref{fig_abs_ref_exs}).

Due to the difficulty of abstractive summarization, the great majority of past work has been extractive \cite{kupiec1995trainable, paice1990constructing, saggion2013automatic}.
However, the recent success of \textit{sequence-to-sequence} models \cite{sutskever2014sequence}, in which recurrent neural networks (RNNs) both read and freely generate text, has made abstractive summarization viable \cite{chopraabstractive,nallapati2016abstractive,rush2015neural,zeng2016efficient}.
Though these systems are promising, they exhibit undesirable behavior such as inaccurately reproducing factual details, an inability to deal with out-of-vocabulary (OOV) words, and repeating themselves (see Figure \ref{fig_firstpage}).

In this paper we present an architecture that addresses these three 
issues in the context of multi-sentence summaries.
While most recent abstractive work has focused on headline generation
tasks (reducing one or two sentences to a single headline), we believe
that longer-text summarization is both more challenging (requiring
higher levels of abstraction while avoiding repetition) and ultimately more useful.
Therefore we apply our model to the recently-introduced
\textit{CNN\slash Daily Mail} dataset \cite{hermann2015teaching, nallapati2016abstractive}, which contains news articles (39 sentences on average) paired with multi-sentence summaries, and show that we outperform the state-of-the-art abstractive system by at least 2 ROUGE points.

Our hybrid \textit{pointer-generator} network facilitates copying
words from the source text via \textit{pointing} \cite{vinyals2015pointer}, which improves accuracy and handling of OOV words, while retaining the ability to \textit{generate} new words.
The network, which can be viewed as a balance between extractive and abstractive approaches, is similar to \citeauthor{gu2016incorporating}'s \shortcite{gu2016incorporating} CopyNet and \citeauthor{miao2016language}'s \shortcite{miao2016language} Forced-Attention Sentence Compression, that were applied to short-text summarization.
We propose a novel variant of the \textit{coverage vector} \cite{tu2016modeling} from Neural Machine Translation, which we use to track and control coverage of the source document.
We show that coverage is remarkably effective for eliminating repetition.
\end{section} 

\begin{section}{Our Models}
In this section we describe (1) our baseline sequence-to-sequence model, (2) our pointer-generator model, and (3) our coverage mechanism that can be added to either of the first two models. The code for our models is available online.\footnote{\url{www.github.com/abisee/pointer-generator}}

\begin{figure*}[t]
\centering
\includegraphics[width=0.9\textwidth, trim={15 30 65 25},clip]{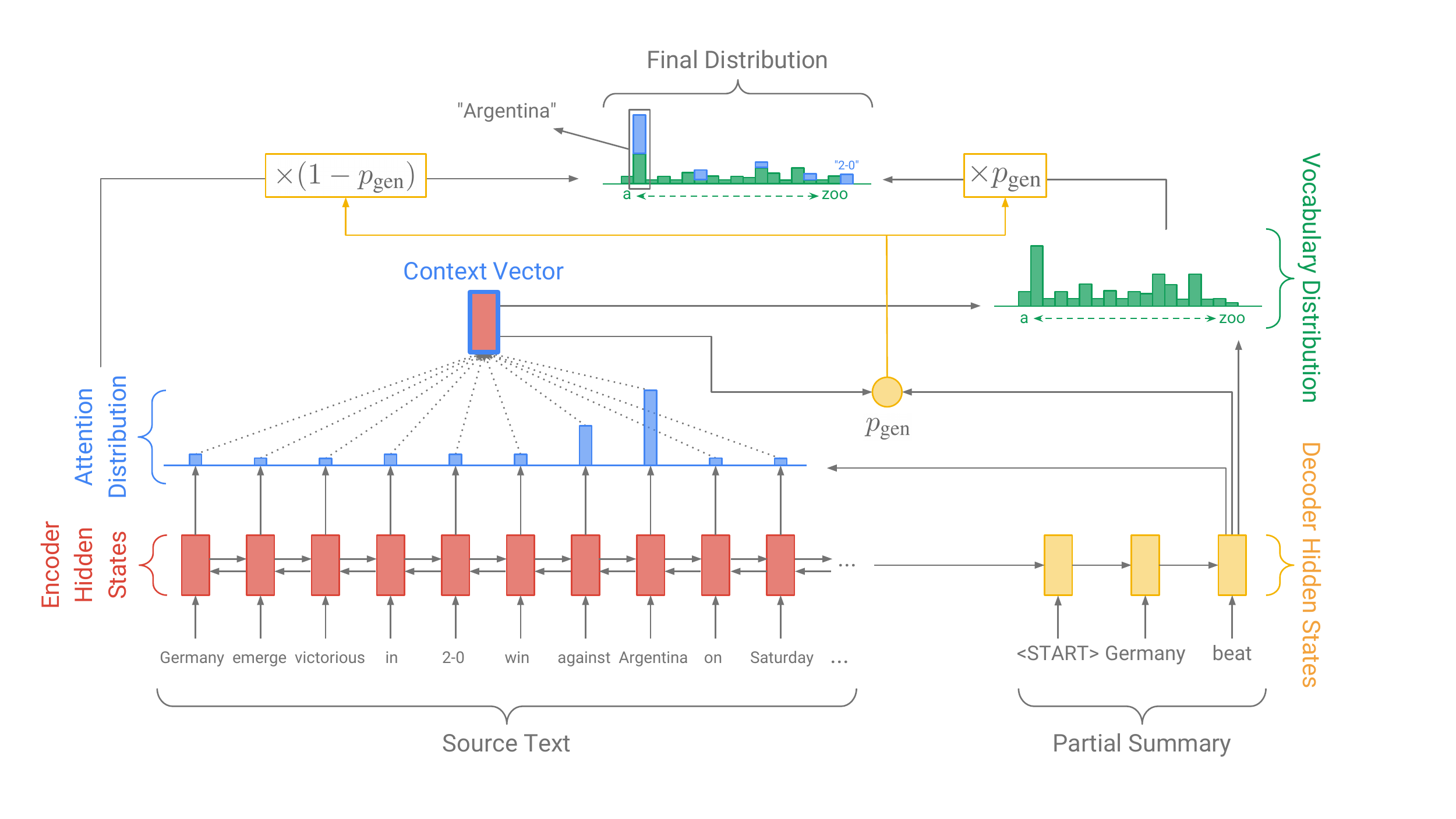} 
\caption{Pointer-generator model.
For each decoder timestep a generation probability $\pgen \in [0,1]$ is calculated, which weights the probability of \textit{generating} words from the vocabulary, versus \textit{copying} words from the source text.
The vocabulary distribution and the attention distribution are weighted and summed to obtain the final distribution, from which we make our prediction.
Note that out-of-vocabulary article words such as \textit{2-0} are included in the final distribution.
Best viewed in color.
}\label{fig_pg_diagram}
\end{figure*}

\begin{subsection}{Sequence-to-sequence attentional model}
\label{subsec_vanilla}
Our baseline model is similar to that of \citeauthor{nallapati2016abstractive} \shortcite{nallapati2016abstractive}, and is depicted in Figure \ref{fig_vanilla_diagram}.
The tokens of the article $w_i$ are fed one-by-one into the encoder (a single-layer bidirectional LSTM), 
producing a sequence of \textit{encoder hidden states} $h_i$.
On each step $t$, the decoder (a single-layer unidirectional LSTM) receives the word embedding of the previous word (while training, this is the previous word of the reference summary; at test time it is the previous word emitted by the decoder), and has \textit{decoder state} $s_t$.
The \textit{attention distribution} $a^t$ is calculated as in \citeauthor{bahdanau2014neural} \shortcite{bahdanau2014neural}:
\begin{align}
e_i^t &= v^T \tanh(W_h h_i + W_s s_t + b_\text{attn})  \label{eqn_vanilla_attn_scores} \\
a^t &= \text{softmax}(e^t) \label{eqn_vanilla_attn_dist}
\end{align}
where $v$, $W_h$, $W_s$ and $b_\text{attn}$ are learnable parameters.
The attention distribution can be viewed as a probability distribution over the source words, that tells the decoder where to look to produce the next word.
Next, the attention distribution is used to produce a weighted sum of the encoder hidden states, known as the \textit{context vector} $h^*_t$:
\begin{equation}
\label{eqn_context_vec}
h^*_t = \sum\nolimits_i a_i^t h_i
\end{equation}
The context vector, which can be seen as a fixed-size representation of what has been read from the source for this step, is concatenated with the decoder state $s_t$ and fed through two linear layers to produce the vocabulary distribution $\pvocab$:
\begin{align}
\pvocab &= \text{softmax}(V' (V [s_t, h^*_t] + b) + b')
\end{align}
where $V$, $V'$, $b$ and $b'$ are learnable parameters.
$\pvocab$ is a probability distribution over all words in the vocabulary, and provides us with our final distribution from which to predict words $w$:
\begin{equation}
P(w) = \pvocab(w)
\end{equation}
During training, the loss for timestep $t$ is the negative log likelihood of the target word $w^*_t$ for that timestep:
\begin{equation}
\label{eqn_vanilla_losst}
\text{loss}_t = -\log P(w^*_t)
\end{equation}
and the overall loss for the whole sequence is:
\begin{equation}
\label{eqn_vanilla_loss}
\text{loss} = \frac{1}{T}\sum\nolimits_{t=0}^T \text{loss}_t
\end{equation}
\end{subsection}

\begin{subsection}{Pointer-generator network}
Our pointer-generator network is a hybrid between our baseline and a pointer network \cite{vinyals2015pointer}, as it allows both copying words via pointing, and generating words from a fixed vocabulary.
In the pointer-generator model (depicted in Figure \ref{fig_pg_diagram}) the attention distribution $a^t$ and context vector $h^*_t$ are calculated as in section \ref{subsec_vanilla}.
In addition, the \textit{generation probability} $\pgen \in [0,1]$ for timestep $t$ is calculated from the context vector $h^*_t$, the decoder state $s_t$ and the decoder input $x_t$:
\begin{equation}
\label{eqn_pgen_calc}
\pgen = \sigma(w_{h^*}^T h^*_t + w_s^T s_t + w_x^T x_t + b_\text{ptr})
\end{equation}
where vectors $w_{h^*}$, $w_s$, $w_x$ and scalar $b_\text{ptr}$ are learnable parameters and $\sigma$ is the sigmoid function.
Next, $\pgen$ is used as a soft switch to choose between \textit{generating} a word from the vocabulary by sampling from $\pvocab$, or \textit{copying} a word from the input sequence by sampling from the attention distribution $a^t$.
For each document let the \textit{extended vocabulary} denote the union of the vocabulary, and all words appearing in the source document.
We obtain the following probability distribution over the extended vocabulary:
\begin{equation}
\label{eqn_pg_dist}
P(w) = \pgen \pvocab(w) + (1-\pgen)\sum\nolimits_{i: w_i = w} a^t_i
\end{equation}
Note that if $w$ is an out-of-vocabulary (OOV) word, then $\pvocab(w)$ is zero;
similarly if $w$ does not appear in the source document, then $\sum_{i: w_i = w} a^t_i$ is zero.
The ability to produce OOV words is one of the primary advantages of pointer-generator models; by contrast models such as our baseline are restricted to their pre-set vocabulary.

The loss function is as described in equations (\ref{eqn_vanilla_losst}) and (\ref{eqn_vanilla_loss}), but with respect to our modified probability distribution $P(w)$ given in equation (\ref{eqn_pg_dist}).
\end{subsection}

\begin{subsection}{Coverage mechanism}
Repetition is a common problem for sequence-to-sequence models \cite{tu2016modeling, mi2016coverage, sankaran2016temporal, suzuki2016rnn}, and is especially pronounced when generating multi-sentence text (see Figure \ref{fig_firstpage}).
We adapt the \textit{coverage model} of \citeauthor{tu2016modeling}  \shortcite{tu2016modeling} to solve the problem.
In our coverage model, we maintain a \textit{coverage vector} $c^t$, which is the sum of attention distributions over all previous decoder timesteps:
\begin{equation}
c^t = \sum\nolimits_{t'=0}^{t-1} a^{t'}
\end{equation}
Intuitively, $c^t$ is a (unnormalized) distribution over the source document words that represents the degree of coverage that those words have received from the attention mechanism so far.
Note that $c^0$ is a zero vector, because on the first timestep, none of the source document has been covered.

The coverage vector is used as extra input to the attention mechanism, changing equation (\ref{eqn_vanilla_attn_scores}) to:
\begin{equation}
\label{eqn_coverage_attn_scores}
e_i^t = v^T \tanh(W_h h_i + W_s s_t + w_c c^t_i + b_\text{attn})
\end{equation}
where $w_c$ is a learnable parameter vector of same length as $v$.
This ensures that the attention mechanism's current decision (choosing where to attend next) is informed by a reminder of its previous decisions (summarized in $c^t$).
This should make it easier for the attention mechanism to avoid repeatedly attending to the same locations, and thus avoid generating repetitive text.

We find it necessary (see section \ref{sec_experiments}) to additionally define a \textit{coverage loss} to penalize repeatedly attending to the same locations:
\begin{equation}
\label{eqn_covloss}
\text{covloss}_t = \sum\nolimits_i \min(a_i^t,c_i^t)
\end{equation}
Note that the coverage loss is bounded; in particular $\text{covloss}_t \le \sum_i a_i^t = 1$.
Equation (\ref{eqn_covloss}) differs from the coverage loss used in Machine Translation.
In MT, we assume that there should be a roughly one-to-one translation ratio; accordingly the final coverage vector is penalized if it is more or less than 1.
Our loss function is more flexible: because summarization should not require uniform coverage, we only penalize the overlap between each attention distribution and the coverage so far -- preventing repeated attention.
Finally, the coverage loss, reweighted by some hyperparameter $\lambda$, is added to the primary loss function to yield a new composite loss function:
\begin{equation}
\label{eqn_coverage_totalloss}
\text{loss}_t = -\log P(w^*_t) + \lambda \sum\nolimits_i \min(a_i^t,c_i^t)
\end{equation}
\end{subsection}
\end{section} 

\begin{section}{Related Work}
\textbf{Neural abstractive summarization.}
\citeauthor{rush2015neural} \shortcite{rush2015neural} were the first to apply modern neural networks to abstractive text summarization, achieving state-of-the-art performance on DUC-2004 and Gigaword, two sentence-level summarization datasets.
Their approach, which is centered on the attention mechanism, has been augmented with recurrent decoders \cite{chopraabstractive}, Abstract Meaning Representations \cite{takase2016neural}, hierarchical networks \cite{nallapati2016abstractive}, variational autoencoders \cite{miao2016language}, and direct optimization of the performance metric \cite{ranzato2015sequence}, further improving performance on those datasets.

However, large-scale datasets for summarization of \textit{longer} text are rare.
\citeauthor{nallapati2016abstractive} \shortcite{nallapati2016abstractive} adapted the DeepMind question-answering dataset \cite{hermann2015teaching} for summarization, resulting in the \textit{CNN\slash Daily Mail} dataset, and provided the first abstractive baselines.
The same authors then published a neural \emph{extractive} approach \cite{nallapati2016summarunner}, which uses hierarchical RNNs to select sentences, and found that it significantly outperformed their abstractive result with respect to the ROUGE metric.
To our knowledge, these are the only two published results on the full dataset.

Prior to modern neural methods, abstractive summarization received less attention than extractive summarization, but \citeauthor{jing2000sentence} \shortcite{jing2000sentence} explored cutting unimportant parts of sentences to create summaries, and \citeauthor{cheung2014unsupervised} \shortcite{cheung2014unsupervised} explore sentence fusion using dependency trees.

\textbf{Pointer-generator networks.}
The pointer network \cite{vinyals2015pointer} is a sequence-to-sequence model that uses the soft attention distribution of \citeauthor{bahdanau2014neural} \shortcite{bahdanau2014neural} to produce an output sequence consisting of elements from the input sequence.
The pointer network has been used to create hybrid approaches for NMT \cite{gulcehre2016pointing}, language modeling \cite{merity2016pointer}, and summarization \cite{gu2016incorporating,gulcehre2016pointing,miao2016language,nallapati2016abstractive,zeng2016efficient}.

Our approach is close to the Forced-Attention Sentence Compression model of \citeauthor{miao2016language} \shortcite{miao2016language} and the CopyNet model of \citeauthor{gu2016incorporating} \shortcite{gu2016incorporating}, with some small differences:
(i) We calculate an explicit switch probability $\pgen$, whereas Gu et al. induce competition through a shared softmax function.
(ii) We recycle the attention distribution to serve as the copy distribution, but Gu et al. use two separate distributions.
(iii) When a word appears multiple times in the source text, we sum probability mass from all corresponding parts of the attention distribution, whereas Miao and Blunsom do not.
Our reasoning is that (i) calculating an explicit $\pgen$ usefully enables us to raise or lower the probability of all generated words or all copy words at once, rather than individually, (ii) the two distributions serve such similar purposes that we find our simpler approach suffices, and (iii) we observe that the pointer mechanism often copies a word while attending to multiple occurrences of it in the source text.

Our approach is considerably different from that of
\citet{gulcehre2016pointing} and \citet{nallapati2016abstractive}.
Those works train their pointer components to activate only for out-of-vocabulary words or named entities (whereas we allow our model to freely learn when to use the pointer), and they do not mix the probabilities from the copy distribution and the vocabulary distribution.
We believe the mixture approach described here is better for abstractive summarization -- in section \ref{sec_results} we show that the copy mechanism is vital for accurately reproducing rare but in-vocabulary words, and in section \ref{subsec_how_abs} we observe that the mixture model enables the language model and copy mechanism to work together to perform abstractive copying.

\textbf{Coverage.}
Originating from Statistical Machine Translation \cite{koehn2009statistical}, coverage was adapted for NMT by \citeauthor{tu2016modeling} \shortcite{tu2016modeling} and \citeauthor{mi2016coverage} \shortcite{mi2016coverage}, who both use a GRU to update the coverage vector each step.
We find that a simpler approach -- summing the attention distributions to obtain the coverage vector -- suffices.
In this respect our approach is similar to \citeauthor{xu2015show} \shortcite{xu2015show}, who apply a coverage-like method to image captioning,
and \citeauthor{chen2016distraction} \shortcite{chen2016distraction}, who also incorporate a coverage mechanism (which they call `distraction') as described in equation (\ref{eqn_coverage_attn_scores}) into neural summarization of longer text.

\textit{Temporal attention} is a related technique that has been applied to NMT \cite{sankaran2016temporal}  and summarization \cite{nallapati2016abstractive}.
In this approach, each attention distribution is divided by the sum of the previous, which effectively dampens repeated attention.
We tried this method but found it too destructive, distorting the signal from the attention mechanism and reducing performance.
We hypothesize that an early intervention method such as coverage is preferable to a post hoc method such as temporal attention -- it is better to \textit{inform} the attention mechanism to help it make better decisions, than to \textit{override} its decisions altogether.
This theory is supported by the large boost that coverage gives our ROUGE scores (see Table \ref{tab_results}), compared to the smaller boost given by temporal attention for the same task \cite{nallapati2016abstractive}.

\end{section} 

\begin{section}{Dataset}
\label{sec_dataset}
We use the \textit{CNN\slash Daily Mail} dataset \cite{hermann2015teaching,nallapati2016abstractive}, which contains online news articles (781 tokens on average) paired with multi-sentence summaries (3.75 sentences or 56 tokens on average).
We used scripts supplied by \citet{nallapati2016abstractive} to obtain the same version of the the data, which has 287,226 training pairs, 13,368 validation pairs and 11,490 test pairs.
Both the dataset's published results \cite{nallapati2016abstractive, nallapati2016summarunner} use the \textit{anonymized} version of the data, which has been pre-processed to replace each named entity, e.g., \textit{The United Nations}, with its own unique identifier for the example pair, e.g., \texttt{@entity5}.
By contrast, we operate directly on the original text (or \textit{non-anonymized} version of the data),\footnote{at \url{www.github.com/abisee/pointer-generator}} which we believe is the favorable problem to solve because it requires no pre-processing.
\end{section}


\begin{table*}[t]
\centering
\begin{tabular}{ | l | c | c | c | c | c |}
\hline
 & \multicolumn{3}{|c|}{ROUGE} & \multicolumn{2}{|c|}{METEOR} \\ \cline{2-6}
& 1 & 2 & L & exact match & + stem/syn/para \\ \hline
abstractive model \cite{nallapati2016abstractive}* & 35.46 & 13.30 & 32.65 & - & - \\
seq-to-seq + attn baseline (150k vocab) & 30.49 & 11.17 & 28.08 & 11.65 & 12.86 \\
seq-to-seq + attn baseline (50k vocab) & 31.33 & 11.81 & 28.83 & 12.03 & 13.20  \\
pointer-generator & 36.44 & 15.66 & 33.42 & 15.35 & 16.65  \\
pointer-generator + coverage & \textbf{39.53} & \textbf{17.28} & \textbf{36.38} & 17.32 & 18.72 \\ \hline
lead-3 baseline (ours) & 40.34 & 17.70 & 36.57 & 20.48 & 22.21 \\
lead-3 baseline \cite{nallapati2016summarunner}* & 39.2$\phantom0$ & 15.7$\phantom0$ & 35.5$\phantom0$ &  -  & - \\
extractive model \cite{nallapati2016summarunner}* & 39.6$\phantom0$ & 16.2$\phantom0$  & 35.3$\phantom0$ & - & - \\ \hline
\end{tabular}
\caption{ROUGE F$_1$ and METEOR scores on the test set.
Models and baselines in the top half are abstractive, while those in the bottom half are extractive.
Those marked with * were trained and evaluated on the anonymized dataset, and so are not strictly comparable to our results on the original text.
All our ROUGE scores have a 95\% confidence interval of at most $\pm 0.25$ as reported by the official ROUGE script.
The METEOR improvement from the 50k baseline to the pointer-generator model, and from the pointer-generator to the pointer-generator+coverage model, were both found to be statistically significant using an approximate randomization test with $p<0.01$.
}
\label{tab_results}
\end{table*}

\begin{section}{Experiments}
\label{sec_experiments}
For all experiments, our model has 256-dimensional hidden states and 128-dimensional word embeddings. 
For the pointer-generator models, we use a vocabulary of 50k words for both source and target -- note that due to the pointer network's ability to handle OOV words, we can use a smaller vocabulary size than \citeauthor{nallapati2016abstractive}'s \shortcite{nallapati2016abstractive} 150k source and 60k target vocabularies.
For the baseline model, we also try a larger vocabulary size of 150k.

Note that the pointer and the coverage mechanism introduce very few additional parameters to the network: for the models with vocabulary size 50k, the baseline model has 21,499,600 parameters, the pointer-generator adds 1153 extra parameters ($w_{h^*}$, $w_s$, $w_x$ and $b_\text{ptr}$ in equation \ref{eqn_pgen_calc}), and coverage adds 512 extra parameters ($w_c$ in equation \ref{eqn_coverage_attn_scores}).

Unlike \citet{nallapati2016abstractive}, we do not pre-train the word embeddings -- they are learned from scratch during training.
We train using Adagrad \cite{duchi2011adaptive} with learning rate 0.15 and an initial accumulator value of 0.1.
(This was found to work best of Stochastic Gradient Descent, Adadelta, Momentum, Adam and RMSProp).
We use gradient clipping with a maximum gradient norm of 2, but do not use any form of regularization.
We use loss on the validation set to implement early stopping.

During training and at test time we truncate the article to 400 tokens and limit the length of the summary to 100 tokens for training and 120 tokens at test time.\footnote{The upper limit of 120 is mostly invisible: the beam search algorithm is self-stopping and almost never reaches the 120th step.}
This is done to expedite training and testing, but we also found that truncating the article can \textit{raise} the performance of the model (see section \ref{subsec_lead3} for more details).
For training, we found it efficient to start with highly-truncated sequences, then raise the maximum length once converged.
We train on a single Tesla K40m GPU with a batch size of 16.
At test time our summaries are produced using beam search with beam size 4.

We trained both our baseline models for about 600,000 iterations (33 epochs) -- this is similar to the 35 epochs required by \citeauthor{nallapati2016abstractive}'s \shortcite{nallapati2016abstractive} best model. 
Training took 4 days and 14 hours for the 50k vocabulary model, and 8 days 21 hours for the 150k vocabulary model.
We found the pointer-generator model quicker to train, requiring less than 230,000 training iterations (12.8 epochs); a total of 3 days and 4 hours.
In particular, the pointer-generator model makes much quicker progress in the early phases of training.
To obtain our final coverage model, we added the coverage mechanism with coverage loss weighted to $\lambda=1$ (as described in equation \ref{eqn_coverage_totalloss}), and trained for a further 3000 iterations (about 2 hours).
In this time the coverage loss converged to about 0.2, down from an initial value of about 0.5.
We also tried a more aggressive value of $\lambda=2$; this reduced coverage loss but increased the primary loss function, thus we did not use it.

We tried training the coverage model without the loss function, hoping that the attention mechanism may learn by itself not to attend repeatedly to the same locations, but we found this to be ineffective, with no discernible reduction in repetition.
We also tried training with coverage from the first iteration rather than as a separate training phase, but found that in the early phase of training, the coverage objective interfered with the main objective, reducing overall performance.
\end{section}

\begin{section}{Results}
\label{sec_results}
\subsection{Preliminaries}
Our results are given in Table \ref{tab_results}.
We evaluate our models with the standard ROUGE metric \cite{lin2004rouge}, reporting the F$_1$ scores for ROUGE-1, ROUGE-2 and ROUGE-L (which respectively measure the word-overlap, bigram-overlap, and longest common sequence between the reference summary and the summary to be evaluated).
We obtain our ROUGE scores using the \texttt{pyrouge} package.\footnote{\url{pypi.python.org/pypi/pyrouge/0.1.3}}
We also evaluate with the METEOR metric \cite{denkowski:lavie:meteor-wmt:2014}, both in exact match mode (rewarding only exact matches between words) and full mode (which additionally rewards matching stems, synonyms and paraphrases).\footnote{\url{www.cs.cmu.edu/~alavie/METEOR}}

In addition to our own models, we also report the lead-3 baseline (which uses the first three
sentences of the article as a summary), and compare to the only existing
abstractive \cite{nallapati2016abstractive} and extractive \cite{nallapati2016summarunner} models on the full dataset.
The output of our models is available online.\footnote{\url{www.github.com/abisee/pointer-generator}}

Given that we generate plain-text summaries but Nallapati et
al. \shortcite{nallapati2016abstractive, nallapati2016summarunner}
generate anonymized summaries (see Section \ref{sec_dataset}), our ROUGE scores are not strictly comparable.
There is evidence to suggest that the original-text dataset may result in higher ROUGE scores in general than the anonymized dataset -- the lead-3 baseline is higher on the former than the latter.
One possible explanation is that multi-word named entities lead to a higher rate of $n$-gram overlap.
Unfortunately, ROUGE is the only available means of comparison with Nallapati et al.'s work.
Nevertheless, given that the disparity in the lead-3 scores is (+1.1 ROUGE-1, +2.0 ROUGE-2, +1.1 ROUGE-L) points respectively, and our best model scores exceed \citeauthor{nallapati2016abstractive} \shortcite{nallapati2016abstractive} by (+4.07 ROUGE-1, +3.98 ROUGE-2, +3.73 ROUGE-L) points, we may estimate that we outperform the only previous abstractive system by at least 2 ROUGE points all-round.

\begin{figure}[t]
\centering

\begin{tikzpicture}
    \begin{axis}[
width=\columnwidth,
height=4cm,
        major x tick style = transparent,
        ybar,
        ymax=30,
        ymin=0,
        bar width=7pt,
        ymajorgrids = true,
        ylabel = {\% that are duplicates},
        	legend cell align=left,
	legend style={at={(0.5,-0.5)},anchor=north,font=\small},
	legend image post style={yscale=1, xscale=3},
        symbolic x coords={
        1-grams,
        2-grams,
        3-grams,
        4-grams,
        sentences,
        },
        x tick label style={rotate=30, anchor=north east, inner sep=0mm}, 
        xtick = data,
        x=36pt, 
        scaled y ticks = false,
    ]

\addplot[style={gred, draw=gred, fill=gred!30, postaction={pattern=north east lines, pattern color=gred}}]
coordinates {
(1-grams, 28.752)
(2-grams, 15.155)
(3-grams, 13.295)
(4-grams, 12.271)
(sentences, 6.65)
};
\addlegendentry{pointer-generator, no coverage}

\addplot[style={draw=gblue,pattern=horizontal lines light blue,pattern color=gblue}]
coordinates {
(1-grams, 21.268)
(2-grams, 03.367)
(3-grams,1.571)
(4-grams, 1.082)
(sentences, 0.124 )
};
\addlegendentry{pointer-generator + coverage}

\addplot[style={ggreen, draw=ggreen, fill=ggreen!30, postaction={pattern=north west lines, pattern color=ggreen}}]
coordinates {
(1-grams,19.543)
(2-grams,1.543)
(3-grams,0.225)
(4-grams,0.049)
(sentences,0)
};
\addlegendentry{reference summaries}

\end{axis}
\end{tikzpicture}
\caption{Coverage eliminates undesirable repetition.
Summaries from our \textcolor{gred}{\textbf{non-coverage model}} contain many duplicated $n$-grams while our \textcolor{gblue}{\textbf{coverage model}} produces a similar number as the \textcolor{ggreen}{\textbf{reference summaries}}.}
\label{fig_repetition}
\end{figure}
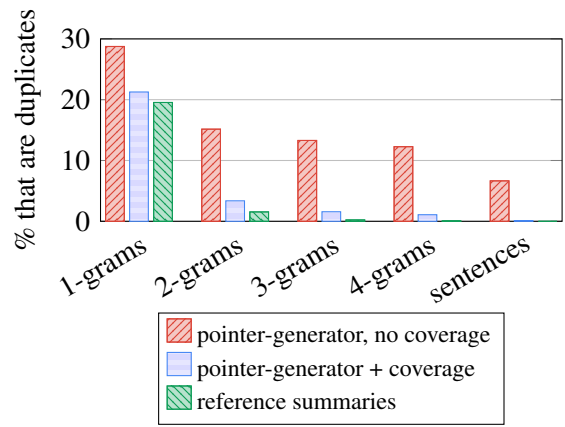

\subsection{Observations}
We find that both our baseline models perform poorly with respect to ROUGE and METEOR, and in fact the larger vocabulary size (150k) does not seem to help.
Even the better-performing baseline (with 50k vocabulary) produces summaries with several common problems.
Factual details are frequently reproduced incorrectly, often replacing an uncommon (but in-vocabulary) word with a more-common alternative.
For example in Figure \ref{fig_firstpage}, the baseline model appears to struggle with the rare word \textit{thwart}, producing \textit{destabilize} instead, which leads to the fabricated phrase \textit{destabilize nigeria's economy}.
Even more catastrophically, the summaries sometimes devolve into repetitive nonsense, such as the third sentence produced by the baseline model in Figure \ref{fig_firstpage}.
In addition, the baseline model can't reproduce out-of-vocabulary words (such as \textit{muhammadu buhari} in Figure \ref{fig_firstpage}).
Further examples of all these problems are provided in the supplementary material.

Our pointer-generator model achieves much better ROUGE and METEOR scores than the baseline, despite many fewer training epochs.
The difference in the summaries is also marked: out-of-vocabulary words are handled easily, factual details are almost always copied correctly, and there are no fabrications (see Figure \ref{fig_firstpage}). 
However, repetition is still very common.

Our pointer-generator model with coverage improves the ROUGE and METEOR scores further, convincingly surpassing the best abstractive model of \citeauthor{nallapati2016abstractive} \shortcite{nallapati2016abstractive} by several ROUGE points.
Despite the brevity of the coverage training phase (about 1\% of the total training time), the repetition problem is almost completely eliminated, which can be seen both qualitatively (Figure \ref{fig_firstpage}) and quantitatively (Figure \ref{fig_repetition}).
However, our best model does not quite surpass the ROUGE scores of the
lead-3 baseline, nor the current best extractive model \cite{nallapati2016summarunner}.
We discuss this issue in section \ref{subsec_lead3}.

\end{section} 

\begin{section}{Discussion}
\label{sec_discussion}

\begin{figure}[t]
\begin{boxedminipage}{\columnwidth}
\small
\textbf{Article:} smugglers lure arab and african migrants by offering discounts to get onto overcrowded ships if people bring more potential passengers, a cnn investigation has revealed. (...) \newline
\textbf{Summary:} cnn investigation \novelh{uncovers} the \novelh{business inside} a \novelh{human smuggling ring}.
\newline \rule{\columnwidth}{0.4pt}
\textbf{Article:} eyewitness video showing white north charleston police officer michael slager shooting to death an unarmed black man has exposed discrepancies in the reports of the first officers on the scene. (...) \newline
\textbf{Summary:} more \novelh{questions than answers emerge} in \novelh{controversial s.c.} police shooting.
\end{boxedminipage}
\caption{Examples of highly abstractive reference summaries (\novelh{bold} denotes novel words).}
\label{fig_abs_ref_exs}
\end{figure}

\begin{subsection}{Comparison with extractive systems}
\label{subsec_lead3}
It is clear from Table \ref{tab_results} that extractive systems tend to achieve higher ROUGE scores than abstractive, and that the extractive lead-3 baseline is extremely strong (even the best extractive system beats it by only a small margin).
We offer two possible explanations for these observations.

Firstly, news articles tend to be structured with the most important information at the start; this partially explains the strength of the lead-3 baseline.
Indeed, we found that using only the first 400 tokens (about 20 sentences) of the article yielded significantly higher ROUGE scores than using the first 800 tokens.

Secondly, the nature of the task and the ROUGE metric make extractive approaches and the lead-3 baseline difficult to beat.
The choice of content for the reference summaries is quite subjective -- sometimes the sentences form a self-contained summary; other times they simply showcase a few interesting details from the article.
Given that the articles contain 39 sentences on average, there are many equally valid ways to choose 3 or 4 highlights in this style.
Abstraction introduces even more options (choice of phrasing), further decreasing the likelihood of matching the reference summary.
For example, \textit{smugglers profit from desperate migrants} is a valid alternative abstractive summary for the first example in Figure \ref{fig_abs_ref_exs}, but it scores 0 ROUGE with respect to the reference summary.
This inflexibility of ROUGE is exacerbated by only having one reference summary, which has been shown to lower ROUGE's reliability compared to multiple reference summaries \cite{lin2004looking}.

Due to the subjectivity of the task and thus the diversity of valid summaries, it seems that ROUGE rewards safe strategies such as selecting the first-appearing content, or preserving original phrasing.
While the reference summaries \textit{do} sometimes deviate from these techniques, those deviations are unpredictable enough that the safer strategy obtains higher ROUGE scores on average.
This may explain why extractive systems tend to obtain higher ROUGE scores than abstractive, and even extractive systems do not significantly exceed the lead-3 baseline.

To explore this issue further, we evaluated our systems with the METEOR metric, which rewards not only exact word matches, but also matching stems, synonyms and paraphrases (from a pre-defined list). 
We observe that all our models receive over 1 METEOR point boost by the inclusion of stem, synonym and paraphrase matching, indicating that they may be performing some abstraction.
However, we again observe that the lead-3 baseline is not surpassed by our models.
It may be that news article style makes the lead-3 baseline very strong with respect to any metric.
We believe that investigating this issue further is an important direction for future work.
\end{subsection}

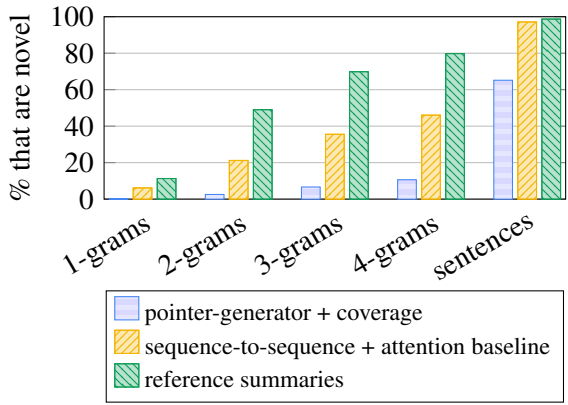
\begin{figure}[t]
\centering

\begin{tikzpicture}
    \begin{axis}[
width=\columnwidth,
height=4cm,
        major x tick style = transparent,
        ybar,
        ymax=100,
        ymin=0,
        bar width=7pt,
        ymajorgrids = true,
        ylabel = {\% that are novel},
        legend cell align=left,
        legend style={at={(0.5,-0.5)},anchor=north,font=\small},
        legend image post style={yscale=1, xscale=3},
        symbolic x coords={
        1-grams,
        2-grams,
        3-grams,
        4-grams,
        sentences,
        },
        x tick label style={rotate=30, anchor=north east, inner sep=0mm}, 
        xtick = data,
        x=36pt, 
        scaled y ticks = false,
    ]

\addplot[style={draw=gblue,pattern=horizontal lines light blue,pattern color=gblue}]
coordinates {
(1-grams,0.068)
(2-grams,2.544)
(3-grams,6.646)
(4-grams,10.627)
(sentences,65.172)
};
\addlegendentry{pointer-generator + coverage}

\addplot[style={gyellow, draw=gyellow, fill=gyellow!30, postaction={pattern=north east lines, pattern color=gyellow}}]
coordinates {
(1-grams,6.151)
(2-grams,21.196)
(3-grams,35.542)
(4-grams,46.05)
(sentences,97.113)
};
\addlegendentry{sequence-to-sequence + attention baseline}

\addplot[style={ggreen, draw=ggreen, fill=ggreen!30, postaction={pattern=north west lines, pattern color=ggreen}}]
coordinates {
(1-grams,11.293)
(2-grams,49.029)
(3-grams,69.846)
(4-grams,79.793)
(sentences,98.715)
};
\addlegendentry{reference summaries}

\end{axis}
\end{tikzpicture}
\caption{Although our \textcolor{gblue}{\textbf{best model}} is abstractive, it does not produce novel $n$-grams (i.e., $n$-grams that don't appear in the source text) as often as the \textcolor{ggreen}{\textbf{reference summaries}}. The \textcolor{gyellow!90!black}{\textbf{baseline model}} produces more novel $n$-grams, but many of these are erroneous (see section \ref{subsec_how_abs}).}
\label{fig_abstractiveness}
\end{figure}

\begin{figure}[h]
\begin{boxedminipage}{\columnwidth}
\small
\textbf{Article:} andy murray (...) is into the semi-finals of the miami open , but not before getting a scare from 21 year-old austrian dominic thiem, who pushed him to 4-4 in the second set before going down 3-6 6-4, 6-1 in an hour and three quarters. (...) \newline
\textbf{Summary:} andy murray \novelh{defeated} dominic thiem 3-6 6-4, 6-1 in an hour and three quarters.
\newline \rule{\columnwidth}{0.4pt}
\textbf{Article:} (...) wayne rooney smashes home during manchester united 's 3-1 win over aston villa on saturday. (...) \newline
\textbf{Summary:} manchester united \novelh{beat} aston villa 3-1 at old trafford on saturday.
\end{boxedminipage}
\caption{Examples of abstractive summaries produced by our model (\novelh{bold} denotes novel words).}
\label{fig_abs_my_exs}
\end{figure}

\begin{subsection}{How abstractive is our model?}
\label{subsec_how_abs}
We have shown that our pointer mechanism makes our abstractive system more reliable, copying factual details correctly more often.
But does the ease of copying make our system any less \textit{abstractive}?

Figure \ref{fig_abstractiveness} shows that our final model's summaries contain a much lower rate of novel $n$-grams (i.e., those that don't appear in the article) than the reference summaries, indicating a lower degree of abstraction.
Note that the baseline model produces novel $n$-grams more frequently -- however, this statistic includes all the incorrectly copied words, \textit{UNK} tokens and fabrications alongside the good instances of abstraction.

In particular, Figure \ref{fig_abstractiveness} shows that our final model copies whole article sentences 35\% of the time; by comparison the reference summaries do so only 1.3\% of the time.
This is a main area for improvement, as we would like our model to move beyond simple sentence extraction.
However, we observe that the other 65\% encompasses a range of abstractive techniques.
Article sentences are truncated to form grammatically-correct shorter versions, and new sentences are composed by stitching together fragments.
Unnecessary interjections, clauses and parenthesized phrases are sometimes omitted from copied passages.
Some of these abilities are demonstrated in Figure \ref{fig_firstpage}, and the supplementary material contains more examples.

Figure \ref{fig_abs_my_exs} shows two examples of more impressive abstraction -- both with similar structure.
The dataset contains many sports stories whose summaries follow the \textit{X beat Y $\langle$score$\rangle$ on $\langle$day$\rangle$} template, which may explain why our model is most confidently abstractive on these examples.
In general however, our model does not routinely produce summaries like those in Figure \ref{fig_abs_my_exs}, and is not close to producing summaries like in Figure \ref{fig_abs_ref_exs}.

The value of the generation probability $\pgen$ also gives a measure of the abstractiveness of our model.
During training, $\pgen$ starts with a value of about 0.30 then increases, converging to about 0.53 by the end of training. 
This indicates that the model first learns to mostly copy, then learns to generate about half the time.
However at test time, $\pgen$ is heavily skewed towards copying, with a mean value of 0.17.
The disparity is likely due to the fact that during training, the model receives word-by-word supervision in the form of the reference summary, but at test time it does not.
Nonetheless, the generator module is useful even when the model is copying.
We find that $\pgen$ is highest at times of uncertainty such as the beginning of sentences, the join between stitched-together fragments, and when producing periods that truncate a copied sentence.
Our mixture model allows the network to copy while simultaneously consulting the language model -- enabling operations like stitching and truncation to be performed with grammaticality. 
In any case, encouraging the pointer-generator model to write more abstractively, while retaining the accuracy advantages of the pointer module, is an exciting direction for future work.
\end{subsection}
\end{section} 

\begin{section}{Conclusion}
In this work we presented a hybrid pointer-generator architecture with coverage, and showed that it reduces inaccuracies and repetition.
We applied our model to a new and challenging long-text dataset, and significantly outperformed the abstractive state-of-the-art result.
Our model exhibits many abstractive abilities, but attaining higher levels of abstraction remains an open research question.
\end{section}

\begin{section}{Acknowledgment}
We thank the ACL reviewers for their helpful comments.
This work was begun while the first author was an intern at Google Brain and continued at Stanford. Stanford University gratefully acknowledges the support of the DARPA DEFT Program AFRL contract no. FA8750-13-2-0040. Any opinions in this material are those of the authors alone.
\end{section}

\clearpage
\newpage
\bibliographystyle{acl_natbib}
\bibliography{main}

\begin{thebibliography}{}
\expandafter\ifx\csname natexlab\endcsname\relax\def\natexlab#1{#1}\fi

\bibitem[{Bahdanau et~al.(2015)Bahdanau, Cho, and Bengio}]{bahdanau2014neural}
Dzmitry Bahdanau, Kyunghyun Cho, and Yoshua Bengio. 2015.
\newblock Neural machine translation by jointly learning to align and
  translate.
\newblock In {\em International Conference on Learning Representations\/}.

\bibitem[{Chen et~al.(2016)Chen, Zhu, Ling, Wei, and
  Jiang}]{chen2016distraction}
Qian Chen, Xiaodan Zhu, Zhenhua Ling, Si~Wei, and Hui Jiang. 2016.
\newblock Distraction-based neural networks for modeling documents.
\newblock In {\em International Joint Conference on Artificial Intelligence\/}.

\bibitem[{Cheung and Penn(2014)}]{cheung2014unsupervised}
Jackie Chi~Kit Cheung and Gerald Penn. 2014.
\newblock Unsupervised sentence enhancement for automatic summarization.
\newblock In {\em Empirical Methods in Natural Language Processing\/}.

\bibitem[{Chopra et~al.(2016)Chopra, Auli, and Rush}]{chopraabstractive}
Sumit Chopra, Michael Auli, and Alexander~M Rush. 2016.
\newblock Abstractive sentence summarization with attentive recurrent neural
  networks.
\newblock In {\em North American Chapter of the Association for Computational
  Linguistics\/}.

\bibitem[{Denkowski and Lavie(2014)}]{denkowski:lavie:meteor-wmt:2014}
Michael Denkowski and Alon Lavie. 2014.
\newblock Meteor universal: Language specific translation evaluation for any
  target language.
\newblock In {\em EACL 2014 Workshop on Statistical Machine Translation\/}.

\bibitem[{Duchi et~al.(2011)Duchi, Hazan, and Singer}]{duchi2011adaptive}
John Duchi, Elad Hazan, and Yoram Singer. 2011.
\newblock Adaptive subgradient methods for online learning and stochastic
  optimization.
\newblock {\em Journal of Machine Learning Research\/} 12:2121--2159.

\bibitem[{Gu et~al.(2016)Gu, Lu, Li, and Li}]{gu2016incorporating}
Jiatao Gu, Zhengdong Lu, Hang Li, and Victor~OK Li. 2016.
\newblock Incorporating copying mechanism in sequence-to-sequence learning.
\newblock In {\em Association for Computational Linguistics\/}.

\bibitem[{Gulcehre et~al.(2016)Gulcehre, Ahn, Nallapati, Zhou, and
  Bengio}]{gulcehre2016pointing}
Caglar Gulcehre, Sungjin Ahn, Ramesh Nallapati, Bowen Zhou, and Yoshua Bengio.
  2016.
\newblock Pointing the unknown words.
\newblock In {\em Association for Computational Linguistics\/}.

\bibitem[{Hermann et~al.(2015)Hermann, Kocisky, Grefenstette, Espeholt, Kay,
  Suleyman, and Blunsom}]{hermann2015teaching}
Karl~Moritz Hermann, Tomas Kocisky, Edward Grefenstette, Lasse Espeholt, Will
  Kay, Mustafa Suleyman, and Phil Blunsom. 2015.
\newblock Teaching machines to read and comprehend.
\newblock In {\em Neural Information Processing Systems\/}.

\bibitem[{Jing(2000)}]{jing2000sentence}
Hongyan Jing. 2000.
\newblock Sentence reduction for automatic text summarization.
\newblock In {\em Applied natural language processing\/}.

\bibitem[{Koehn(2009)}]{koehn2009statistical}
Philipp Koehn. 2009.
\newblock {\em Statistical machine translation\/}.
\newblock Cambridge University Press.

\bibitem[{Kupiec et~al.(1995)Kupiec, Pedersen, and Chen}]{kupiec1995trainable}
Julian Kupiec, Jan Pedersen, and Francine Chen. 1995.
\newblock A trainable document summarizer.
\newblock In {\em International ACM SIGIR conference on Research and
  development in information retrieval\/}.

\bibitem[{Lin(2004{\natexlab{a}})}]{lin2004looking}
Chin-Yew Lin. 2004{\natexlab{a}}.
\newblock Looking for a few good metrics: Automatic summarization
  evaluation-how many samples are enough?
\newblock In {\em NACSIS/NII Test Collection for Information Retrieval (NTCIR)
  Workshop\/}.

\bibitem[{Lin(2004{\natexlab{b}})}]{lin2004rouge}
Chin-Yew Lin. 2004{\natexlab{b}}.
\newblock Rouge: A package for automatic evaluation of summaries.
\newblock In {\em Text summarization branches out: ACL workshop\/}.

\bibitem[{Merity et~al.(2016)Merity, Xiong, Bradbury, and
  Socher}]{merity2016pointer}
Stephen Merity, Caiming Xiong, James Bradbury, and Richard Socher. 2016.
\newblock Pointer sentinel mixture models.
\newblock In {\em NIPS 2016 Workshop on Multi-class and Multi-label Learning in
  Extremely Large Label Spaces\/}.

\bibitem[{Mi et~al.(2016)Mi, Sankaran, Wang, and Ittycheriah}]{mi2016coverage}
Haitao Mi, Baskaran Sankaran, Zhiguo Wang, and Abe Ittycheriah. 2016.
\newblock Coverage embedding models for neural machine translation.
\newblock In {\em Empirical Methods in Natural Language Processing\/}.

\bibitem[{Miao and Blunsom(2016)}]{miao2016language}
Yishu Miao and Phil Blunsom. 2016.
\newblock Language as a latent variable: Discrete generative models for
  sentence compression.
\newblock In {\em Empirical Methods in Natural Language Processing\/}.

\bibitem[{Nallapati et~al.(2017)Nallapati, Zhai, and
  Zhou}]{nallapati2016summarunner}
Ramesh Nallapati, Feifei Zhai, and Bowen Zhou. 2017.
\newblock {SummaRuNNer}: A recurrent neural network based sequence model for
  extractive summarization of documents.
\newblock In {\em Association for the Advancement of Artificial
  Intelligence\/}.

\bibitem[{Nallapati et~al.(2016)Nallapati, Zhou, dos Santos, Gul{\c{c}}ehre,
  and Xiang}]{nallapati2016abstractive}
Ramesh Nallapati, Bowen Zhou, Cicero dos Santos, {\c{C}}aglar Gul{\c{c}}ehre,
  and Bing Xiang. 2016.
\newblock Abstractive text summarization using sequence-to-sequence {RNN}s and
  beyond.
\newblock In {\em Computational Natural Language Learning\/}.

\bibitem[{Paice(1990)}]{paice1990constructing}
Chris~D Paice. 1990.
\newblock Constructing literature abstracts by computer: techniques and
  prospects.
\newblock {\em Information Processing \& Management\/} 26(1):171--186.

\bibitem[{Ranzato et~al.(2016)Ranzato, Chopra, Auli, and
  Zaremba}]{ranzato2015sequence}
Marc'Aurelio Ranzato, Sumit Chopra, Michael Auli, and Wojciech Zaremba. 2016.
\newblock Sequence level training with recurrent neural networks.
\newblock In {\em International Conference on Learning Representations\/}.

\bibitem[{Rush et~al.(2015)Rush, Chopra, and Weston}]{rush2015neural}
Alexander~M Rush, Sumit Chopra, and Jason Weston. 2015.
\newblock A neural attention model for abstractive sentence summarization.
\newblock In {\em Empirical Methods in Natural Language Processing\/}.

\bibitem[{Saggion and Poibeau(2013)}]{saggion2013automatic}
Horacio Saggion and Thierry Poibeau. 2013.
\newblock Automatic text summarization: Past, present and future.
\newblock In {\em Multi-source, Multilingual Information Extraction and
  Summarization\/}, Springer, pages 3--21.

\bibitem[{Sankaran et~al.(2016)Sankaran, Mi, Al-Onaizan, and
  Ittycheriah}]{sankaran2016temporal}
Baskaran Sankaran, Haitao Mi, Yaser Al-Onaizan, and Abe Ittycheriah. 2016.
\newblock Temporal attention model for neural machine translation.
\newblock {\em arXiv preprint arXiv:1608.02927\/} .

\bibitem[{Sutskever et~al.(2014)Sutskever, Vinyals, and
  Le}]{sutskever2014sequence}
Ilya Sutskever, Oriol Vinyals, and Quoc~V Le. 2014.
\newblock Sequence to sequence learning with neural networks.
\newblock In {\em Neural Information Processing Systems\/}.

\bibitem[{Suzuki and Nagata(2016)}]{suzuki2016rnn}
Jun Suzuki and Masaaki Nagata. 2016.
\newblock {RNN}-based encoder-decoder approach with word frequency estimation.
\newblock {\em arXiv preprint arXiv:1701.00138\/} .

\bibitem[{Takase et~al.(2016)Takase, Suzuki, Okazaki, Hirao, and
  Nagata}]{takase2016neural}
Sho Takase, Jun Suzuki, Naoaki Okazaki, Tsutomu Hirao, and Masaaki Nagata.
  2016.
\newblock Neural headline generation on abstract meaning representation.
\newblock In {\em Empirical Methods in Natural Language Processing\/}.

\bibitem[{Tu et~al.(2016)Tu, Lu, Liu, Liu, and Li}]{tu2016modeling}
Zhaopeng Tu, Zhengdong Lu, Yang Liu, Xiaohua Liu, and Hang Li. 2016.
\newblock Modeling coverage for neural machine translation.
\newblock In {\em Association for Computational Linguistics\/}.

\bibitem[{Vinyals et~al.(2015)Vinyals, Fortunato, and
  Jaitly}]{vinyals2015pointer}
Oriol Vinyals, Meire Fortunato, and Navdeep Jaitly. 2015.
\newblock Pointer networks.
\newblock In {\em Neural Information Processing Systems\/}.

\bibitem[{Xu et~al.(2015)Xu, Ba, Kiros, Cho, Courville, Salakhutdinov, Zemel,
  and Bengio}]{xu2015show}
Kelvin Xu, Jimmy Ba, Ryan Kiros, Kyunghyun Cho, Aaron~C Courville, Ruslan
  Salakhutdinov, Richard~S Zemel, and Yoshua Bengio. 2015.
\newblock Show, attend and tell: Neural image caption generation with visual
  attention.
\newblock In {\em International Conference on Machine Learning\/}.

\bibitem[{Zeng et~al.(2016)Zeng, Luo, Fidler, and Urtasun}]{zeng2016efficient}
Wenyuan Zeng, Wenjie Luo, Sanja Fidler, and Raquel Urtasun. 2016.
\newblock Efficient summarization with read-again and copy mechanism.
\newblock {\em arXiv preprint arXiv:1611.03382\/} .

\end{thebibliography}

\clearpage
\newpage

\appendix
\onecolumn
\section*{Supplementary Material}
This appendix provides examples from the test set, with side-by-side comparisons of the reference summaries and the summaries produced by our models.
In each example:
\begin{itemize}
\item \textit{italics} denote out-of-vocabulary words
\item \error{red} denotes factual errors in the summaries
\item \hlpgen{green shading}{30} intensity represents the value of the generation probability $\pgen$
\item \hlcov{yellow shading}{30} intensity represents final value of the coverage vector at the end of final model's summarization process.
\end{itemize}

\begin{figure*}
\begin{boxedminipage}{\textwidth}
\textbf{Article (truncated):}  \hlcov{andy}{31.87} \hlcov{murray}{13.79} \hlcov{came}{5.57} \hlcov{close}{0.02} \hlcov{to}{0.00} \hlcov{giving}{0.14} \hlcov{himself}{0.00} \hlcov{some}{0.00} \hlcov{extra}{0.00} \hlcov{preparation}{0.00} \hlcov{time}{0.00} \hlcov{for}{0.00} \hlcov{his}{0.05} \hlcov{wedding}{0.05} \hlcov{next}{0.02} \hlcov{week}{0.01} \hlcov{before}{0.00} \hlcov{ensuring}{0.00} \hlcov{that}{0.00} \hlcov{he}{0.04} \hlcov{still}{0.03} \hlcov{has}{0.01} \hlcov{unfinished}{0.00} \hlcov{tennis}{0.01} \hlcov{business}{0.00} \hlcov{to}{0.00} \hlcov{attend}{0.00} \hlcov{to}{0.00} \hlcov{.}{0.00} \hlcov{the}{0.75} \hlcov{world}{2.86} \hlcov{no}{0.11} \hlcov{4}{0.00} \hlcov{is}{0.05} \hlcov{into}{0.02} \hlcov{the}{0.19} \hlcov{semi-finals}{1.17} \hlcov{of}{0.08} \hlcov{the}{5.42} \hlcov{miami}{1.78} \hlcov{open}{0.03} \hlcov{,}{0.01} \hlcov{but}{1.01} \hlcov{not}{0.16} \hlcov{before}{0.01} \hlcov{getting}{0.03} \hlcov{a}{0.77} \hlcov{scare}{0.22} \hlcov{from}{0.03} \hlcov{21}{3.05} \hlcov{year-old}{0.04} \hlcov{austrian}{4.26} \hlcov{dominic}{0.15} \hlcov{\textit{thiem}}{1.40} \hlcov{,}{0.06} \hlcov{who}{0.00} \hlcov{pushed}{0.88} \hlcov{him}{0.00} \hlcov{to}{0.07} \hlcov{4-4}{0.02} \hlcov{in}{0.02} \hlcov{the}{1.13} \hlcov{second}{3.28} \hlcov{set}{1.14} \hlcov{before}{0.06} \hlcov{going}{0.03} \hlcov{down}{0.03} \hlcov{3-6}{22.48} \hlcov{6-4}{34.08} \hlcov{,}{30.71} \hlcov{6-1}{8.99} \hlcov{in}{35.97} \hlcov{an}{33.37} \hlcov{hour}{34.06} \hlcov{and}{35.37} \hlcov{three}{36.15} \hlcov{quarters}{36.12} \hlcov{.}{36.27} \hlcov{murray}{22.98} \hlcov{was}{31.14} \hlcov{awaiting}{37.47} \hlcov{the}{34.46} \hlcov{winner}{34.81} \hlcov{from}{36.18} \hlcov{the}{31.21} \hlcov{last}{36.21} \hlcov{eight}{36.36} \hlcov{match}{35.56} \hlcov{between}{36.30} \hlcov{tomas}{40.00} \hlcov{berdych}{37.90} \hlcov{and}{36.14} \hlcov{argentina}{35.99} \hlcov{'s}{36.13} \hlcov{juan}{36.94} \hlcov{monaco}{36.22} \hlcov{.}{36.13} \hlcov{prior}{14.89} \hlcov{to}{36.12} \hlcov{this}{35.18} \hlcov{tournament}{35.96} \hlcov{\textit{thiem}}{36.80} \hlcov{lost}{35.24} \hlcov{in}{35.56} \hlcov{the}{29.34} \hlcov{second}{31.58} \hlcov{round}{35.11} \hlcov{of}{34.73} \hlcov{a}{35.09} \hlcov{challenger}{37.02} \hlcov{event}{36.14} \hlcov{to}{36.99} \hlcov{soon-to-be}{35.93} \hlcov{new}{36.67} \hlcov{brit}{36.54} \hlcov{\textit{aljaz}}{37.01} \hlcov{bedene}{37.87} \hlcov{.}{36.16} \hlcov{andy}{1.64} \hlcov{murray}{24.45} \hlcov{pumps}{29.24} \hlcov{his}{0.02} \hlcov{first}{0.02} \hlcov{after}{0.08} \hlcov{defeating}{0.09} \hlcov{dominic}{30.22} \hlcov{\textit{thiem}}{34.76} \hlcov{to}{10.67} \hlcov{reach}{0.02} \hlcov{the}{0.90} \hlcov{miami}{1.66} \hlcov{open}{0.02} \hlcov{semi}{1.44} \hlcov{finals}{0.00} \hlcov{.}{0.21} \hlcov{\textit{muray}}{10.29} \hlcov{throws}{0.08} \hlcov{his}{0.01} \hlcov{\textit{sweatband}}{0.02} \hlcov{into}{0.00} \hlcov{the}{0.12} \hlcov{crowd}{0.14} \hlcov{after}{0.04} \hlcov{completing}{0.03} \hlcov{a}{0.05} \hlcov{3-6}{1.61} \hlcov{,}{2.25} \hlcov{6-4}{3.43} \hlcov{,}{5.48} \hlcov{6-1}{25.15} \hlcov{victory}{0.01} \hlcov{in}{0.01} \hlcov{florida}{0.20} \hlcov{.}{0.01} \hlcov{murray}{9.21} \hlcov{shakes}{0.09} \hlcov{hands}{0.00} \hlcov{with}{0.00} \hlcov{\textit{thiem}}{0.02} \hlcov{who}{0.00} \hlcov{he}{0.04} \hlcov{described}{0.06} \hlcov{as}{0.00} \hlcov{a}{0.00} \hlcov{'}{0.01} \hlcov{strong}{0.00} \hlcov{guy}{0.00} \hlcov{'}{0.00} \hlcov{after}{0.00} \hlcov{the}{0.04} \hlcov{game}{0.02} \hlcov{.}{0.29} \hlcov{and}{0.60} \hlcov{murray}{11.69} \hlcov{has}{2.09} \hlcov{a}{0.00} \hlcov{fairly}{0.00} \hlcov{simple}{0.00} \hlcov{message}{0.00} \hlcov{for}{0.00} \hlcov{any}{0.00} \hlcov{of}{0.04} \hlcov{his}{0.02} \hlcov{fellow}{0.00} \hlcov{british}{0.07} \hlcov{tennis}{0.02} \hlcov{players}{0.00} \hlcov{who}{0.00} \hlcov{might}{0.00} \hlcov{be}{0.00} \hlcov{agitated}{0.01} \hlcov{about}{0.00} \hlcov{his}{0.00} \hlcov{imminent}{0.00} \hlcov{arrival}{0.00} \hlcov{into}{0.00} \hlcov{the}{0.04} \hlcov{home}{0.05} \hlcov{ranks}{0.00} \hlcov{:}{0.00} \hlcov{do}{0.03} \hlcov{n't}{0.00} \hlcov{complain}{0.00} \hlcov{.}{0.00} \hlcov{instead}{11.62} \hlcov{the}{0.34} \hlcov{british}{0.88} \hlcov{no}{0.02} \hlcov{1}{0.00} \hlcov{believes}{0.07} \hlcov{his}{0.01} \hlcov{colleagues}{0.00} \hlcov{should}{0.00} \hlcov{use}{0.00} \hlcov{the}{0.00} \hlcov{assimilation}{0.01} \hlcov{of}{0.00} \hlcov{the}{0.02} \hlcov{world}{0.05} \hlcov{number}{0.00} \hlcov{83}{0.00} \hlcov{,}{0.00} \hlcov{originally}{0.01} \hlcov{from}{0.00} \hlcov{slovenia}{0.66} \hlcov{,}{0.00} \hlcov{as}{0.00} \hlcov{motivation}{0.00} \hlcov{to}{0.00} \hlcov{better}{0.00} \hlcov{themselves}{0.00} \hlcov{.}{0.00}
\newline \rule{\columnwidth}{0.4pt}
\textbf{Reference Summary:} \newline british no 1 defeated dominic \textit{thiem} in miami open quarter finals .\newline andy murray celebrated his 500th career win in the previous round .\newline third seed will play the winner of tomas berdych and juan monaco in the semi finals of the atp masters 1000 event in key biscayne .
\newline \rule{\columnwidth}{0.4pt}
\textbf{Baseline:} \newline
andy murray beat UNK \error{bedene 6-3} , 6-4 , 6-1 in an hour and three quarters . \newline
british no 1 believes his colleagues should use the \error{maze} of the world number 83 , originally from slovenia , as motivation to better themselves .
\newline \rule{\columnwidth}{0.4pt}
\textbf{Pointer-Generator, No Coverage:} \newline
andy murray came close to giving himself some extra preparation for his wedding next week before getting a scare from 21 year-old austrian dominic \textit{thiem} . \newline
the world no 4 is into the semi-finals of the miami open , but not before getting a scare from 21 year-old austrian dominic \textit{thiem} .
\newline \rule{\columnwidth}{0.4pt}
\textbf{Pointer-Generator, With Coverage:} \newline  \hlpgen{andy}{2.99} \hlpgen{murray}{1.23} \hlpgen{defeated}{25.22} \hlpgen{dominic}{2.75} \hlpgen{\textit{thiem}}{0.25} \hlpgen{3-6}{7.13} \hlpgen{6-4}{2.60} \hlpgen{,}{8.31} \hlpgen{6-1}{2.31} \hlpgen{in}{25.13} \hlpgen{an}{3.65} \hlpgen{hour}{0.59} \hlpgen{and}{13.80} \hlpgen{three}{3.03} \hlpgen{quarters}{2.65} \hlpgen{.}{33.20}\newline \hlpgen{murray}{5.26} \hlpgen{was}{23.39} \hlpgen{awaiting}{13.55} \hlpgen{the}{3.39} \hlpgen{winner}{4.03} \hlpgen{from}{14.22} \hlpgen{the}{2.00} \hlpgen{last}{1.91} \hlpgen{eight}{1.74} \hlpgen{match}{3.07} \hlpgen{between}{7.15} \hlpgen{tomas}{1.06} \hlpgen{berdych}{0.83} \hlpgen{and}{16.43} \hlpgen{argentina}{0.21} \hlpgen{'s}{3.96} \hlpgen{juan}{0.19} \hlpgen{monaco}{0.05} \hlpgen{.}{9.60}\newline \hlpgen{prior}{15.66} \hlpgen{to}{12.27} \hlpgen{this}{4.35} \hlpgen{tournament}{2.37} \hlpgen{thiem}{1.33} \hlpgen{lost}{0.88} \hlpgen{in}{1.46} \hlpgen{the}{0.35} \hlpgen{second}{0.34} \hlpgen{round}{0.50} \hlpgen{of}{2.75} \hlpgen{a}{1.21} \hlpgen{challenger}{0.68} \hlpgen{event}{0.75} \hlpgen{to}{3.09} \hlpgen{soon-to-be}{1.52} \hlpgen{new}{0.80} \hlpgen{brit}{0.71} \hlpgen{\textit{aljaz}}{0.35} \hlpgen{bedene}{0.09} \hlpgen{.}{4.48}
\end{boxedminipage}
\caption{
The baseline model reports the wrong score \textit{6-3}, substitutes \textit{bedene} for \textit{thiem} and struggles with the uncommon word \textit{assimilation}.
The pointer-network models accurately reproduce the out-of-vocabulary words \textit{thiem} and \textit{aljaz}.
Note that the final model produces the novel word \textit{defeated} to incorporate several fragments into a single sentence.
}
\label{fig_murray}
\end{figure*}

\begin{figure*}
\begin{boxedminipage}{\textwidth}
\textbf{Article (truncated):}  \hlcov{munster}{16.20} \hlcov{have}{0.14} \hlcov{signed}{1.77} \hlcov{new}{7.60} \hlcov{zealand}{0.53} \hlcov{international}{1.18} \hlcov{francis}{15.76} \hlcov{\textit{saili}}{12.63} \hlcov{on}{3.35} \hlcov{a}{3.16} \hlcov{two-year}{1.83} \hlcov{deal}{4.41} \hlcov{.}{0.21} \hlcov{utility}{10.36} \hlcov{back}{0.11} \hlcov{\textit{saili}}{0.69} \hlcov{,}{1.47} \hlcov{who}{0.11} \hlcov{made}{8.29} \hlcov{his}{0.13} \hlcov{all}{0.11} \hlcov{blacks}{0.01} \hlcov{debut}{0.03} \hlcov{against}{0.04} \hlcov{argentina}{0.37} \hlcov{in}{0.15} \hlcov{2013}{10.62} \hlcov{,}{0.14} \hlcov{will}{11.95} \hlcov{move}{0.75} \hlcov{to}{0.05} \hlcov{the}{0.06} \hlcov{province}{0.15} \hlcov{later}{4.25} \hlcov{this}{3.00} \hlcov{year}{2.18} \hlcov{after}{0.07} \hlcov{the}{0.23} \hlcov{completion}{0.18} \hlcov{of}{0.00} \hlcov{his}{0.79} \hlcov{2015}{0.71} \hlcov{contractual}{0.17} \hlcov{commitments}{0.05} \hlcov{.}{0.01} \hlcov{the}{21.43} \hlcov{24-year-old}{34.32} \hlcov{currently}{7.23} \hlcov{plays}{2.30} \hlcov{for}{0.01} \hlcov{\textit{auckland-based}}{0.04} \hlcov{super}{0.27} \hlcov{rugby}{0.06} \hlcov{side}{0.02} \hlcov{the}{0.31} \hlcov{blues}{0.96} \hlcov{and}{0.05} \hlcov{was}{6.40} \hlcov{part}{27.86} \hlcov{of}{30.52} \hlcov{the}{29.19} \hlcov{new}{28.30} \hlcov{zealand}{36.31} \hlcov{under-20}{36.36} \hlcov{side}{37.39} \hlcov{that}{36.36} \hlcov{won}{34.86} \hlcov{the}{25.00} \hlcov{junior}{35.55} \hlcov{world}{36.23} \hlcov{championship}{37.83} \hlcov{in}{25.28} \hlcov{italy}{32.07} \hlcov{in}{37.06} \hlcov{2011}{21.96} \hlcov{.}{35.09} \hlcov{\textit{saili}}{12.08} \hlcov{'s}{25.68} \hlcov{signature}{36.54} \hlcov{is}{37.31} \hlcov{something}{37.37} \hlcov{of}{37.14} \hlcov{a}{37.32} \hlcov{coup}{37.51} \hlcov{for}{37.40} \hlcov{munster}{37.45} \hlcov{and}{37.82} \hlcov{head}{35.97} \hlcov{coach}{37.48} \hlcov{anthony}{40.00} \hlcov{foley}{37.55} \hlcov{believes}{35.42} \hlcov{he}{0.21} \hlcov{will}{0.07} \hlcov{be}{0.01} \hlcov{a}{0.00} \hlcov{great}{0.01} \hlcov{addition}{0.00} \hlcov{to}{0.00} \hlcov{their}{0.01} \hlcov{backline}{0.01} \hlcov{.}{0.33} \hlcov{francis}{6.25} \hlcov{\textit{saili}}{21.57} \hlcov{has}{23.53} \hlcov{signed}{35.35} \hlcov{a}{30.35} \hlcov{two-year}{37.75} \hlcov{deal}{32.00} \hlcov{to}{37.24} \hlcov{join}{36.14} \hlcov{munster}{36.45} \hlcov{and}{7.79} \hlcov{will}{0.97} \hlcov{link}{0.23} \hlcov{up}{0.00} \hlcov{with}{0.01} \hlcov{them}{0.22} \hlcov{later}{22.93} \hlcov{this}{34.73} \hlcov{year}{35.22} \hlcov{.}{37.21} \hlcov{'}{0.36} \hlcov{we}{0.10} \hlcov{are}{0.00} \hlcov{really}{0.00} \hlcov{pleased}{0.01} \hlcov{that}{0.00} \hlcov{francis}{1.45} \hlcov{has}{0.24} \hlcov{committed}{0.03} \hlcov{his}{0.00} \hlcov{future}{0.00} \hlcov{to}{0.00} \hlcov{the}{0.02} \hlcov{province}{0.03} \hlcov{,}{0.03} \hlcov{'}{0.05} \hlcov{foley}{11.73} \hlcov{told}{0.12} \hlcov{munster}{0.13} \hlcov{'s}{0.02} \hlcov{official}{0.03} \hlcov{website}{0.00} \hlcov{.}{2.43} \hlcov{'}{0.46} \hlcov{he}{3.15} \hlcov{is}{0.17} \hlcov{a}{0.02} \hlcov{talented}{0.02} \hlcov{centre}{0.01} \hlcov{with}{0.00} \hlcov{an}{0.03} \hlcov{impressive}{0.00} \hlcov{\textit{skill-set}}{0.06} \hlcov{and}{0.00} \hlcov{he}{0.21} \hlcov{possesses}{0.19} \hlcov{the}{0.01} \hlcov{physical}{0.01} \hlcov{attributes}{0.00} \hlcov{to}{0.00} \hlcov{excel}{0.02} \hlcov{in}{0.00} \hlcov{the}{0.06} \hlcov{northern}{0.00} \hlcov{hemisphere}{0.00} \hlcov{.}{0.06} \hlcov{'}{0.58} \hlcov{i}{0.33} \hlcov{believe}{0.04} \hlcov{he}{0.01} \hlcov{will}{0.02} \hlcov{be}{0.00} \hlcov{a}{0.00} \hlcov{great}{0.01} \hlcov{addition}{0.00} \hlcov{to}{0.00} \hlcov{our}{0.01} \hlcov{backline}{0.01} \hlcov{and}{0.00} \hlcov{we}{0.00} \hlcov{look}{0.00} \hlcov{forward}{0.20} \hlcov{to}{0.00} \hlcov{welcoming}{0.02} \hlcov{him}{0.05} \hlcov{to}{0.00} \hlcov{munster}{0.08} \hlcov{.}{1.86} \hlcov{'}{0.33} \hlcov{\textit{saili}}{24.99} \hlcov{has}{14.19} \hlcov{been}{1.29} \hlcov{capped}{0.63} \hlcov{twice}{0.02} \hlcov{by}{0.00} \hlcov{new}{0.19} \hlcov{zealand}{0.49} \hlcov{and}{0.10} \hlcov{was}{2.47} \hlcov{part}{8.50} \hlcov{of}{6.87} \hlcov{the}{6.43} \hlcov{under}{5.19} \hlcov{20}{0.02} \hlcov{side}{0.02} \hlcov{that}{0.39} \hlcov{won}{2.89} \hlcov{the}{8.92} \hlcov{junior}{4.70} \hlcov{championship}{6.23} \hlcov{in}{12.46} \hlcov{2011}{10.04} \hlcov{.}{2.45}
\newline \rule{\columnwidth}{0.4pt}
\textbf{Reference Summary:} \newline utility back francis \textit{saili} will join up with munster later this year .\newline the new zealand international has signed a two-year contract .\newline \textit{saili} made his debut for the all blacks against argentina in 2013 .
\newline \rule{\columnwidth}{0.4pt}
\textbf{Baseline:} \newline
\error{dutch} international francis UNK has signed a two-year deal to join \error{irish} UNK super rugby side the blues .\newline
UNK 's signature is something of a coup for munster and his head coach anthony foley believes he will be a great addition to their \error{respective prospects} . \newline
UNK has been capped twice by new zealand .
\newline \rule{\columnwidth}{0.4pt}
\textbf{Pointer-Generator, No Coverage:} \newline
new zealand international francis \textit{saili} will move to the province later this year .\newline
utility back \textit{saili} made his all blacks debut against argentina in 2013 .\newline
utility back \textit{saili} will move to the province later this year .
\newline \rule{\columnwidth}{0.4pt}
\textbf{Pointer-Generator, With Coverage:} \newline  \hlpgen{francis}{2.26} \hlpgen{\textit{saili}}{3.74} \hlpgen{has}{25.14} \hlpgen{signed}{13.17} \hlpgen{a}{5.50} \hlpgen{two-year}{1.39} \hlpgen{deal}{3.11} \hlpgen{to}{13.47} \hlpgen{join}{5.56} \hlpgen{munster}{7.87} \hlpgen{later}{28.65} \hlpgen{this}{23.36} \hlpgen{year}{13.92} \hlpgen{.}{39.56}\newline \hlpgen{the}{28.80} \hlpgen{24-year-old}{2.09} \hlpgen{was}{12.26} \hlpgen{part}{11.01} \hlpgen{of}{25.41} \hlpgen{the}{9.59} \hlpgen{new}{4.22} \hlpgen{zealand}{0.71} \hlpgen{under-20}{14.45} \hlpgen{side}{5.96} \hlpgen{that}{15.18} \hlpgen{won}{13.92} \hlpgen{the}{4.10} \hlpgen{junior}{2.88} \hlpgen{world}{1.37} \hlpgen{championship}{1.10} \hlpgen{in}{13.76} \hlpgen{italy}{0.74} \hlpgen{in}{19.16} \hlpgen{2011}{1.14} \hlpgen{.}{29.45}\newline \hlpgen{\textit{saili}}{12.39} \hlpgen{'s}{13.66} \hlpgen{signature}{2.94} \hlpgen{is}{4.60} \hlpgen{something}{1.49} \hlpgen{of}{0.98} \hlpgen{a}{0.81} \hlpgen{coup}{0.30} \hlpgen{for}{3.42} \hlpgen{munster}{1.23} \hlpgen{and}{6.61} \hlpgen{head}{0.62} \hlpgen{coach}{0.55} \hlpgen{anthony}{2.07} \hlpgen{foley}{0.09} \hlpgen{.}{15.54}\newline
\end{boxedminipage}
\caption{The baseline model incorrectly substitutes \textit{dutch} for \textit{new zealand} (perhaps reflecting the European bias of the dataset), fabricates \textit{irish}, and struggles with out-of-vocabulary words \textit{saili} and \textit{auckland-based}.
Though it is not clear why, the phrase \textit{addition  to our backline} is changed to the nonsensical \textit{addition to their respective prospects}.
The pointer-generator model fixes these accuracy problems, and the addition of coverage fixes the repetition problem.
Note that the final model skips over large passages of text to produce shorter sentences.}
\label{fig_saili}
\end{figure*}

\begin{figure*}
\begin{boxedminipage}{\textwidth}
\textbf{Article (truncated):}  \hlcov{right}{0.03} \hlcov{from}{0.00} \hlcov{the}{0.39} \hlcov{moment}{0.01} \hlcov{he}{0.36} \hlcov{breezed}{0.02} \hlcov{through}{0.04} \hlcov{the}{0.07} \hlcov{doors}{0.01} \hlcov{at}{0.44} \hlcov{old}{23.52} \hlcov{trafford}{35.44} \hlcov{,}{10.83} \hlcov{louis}{26.08} \hlcov{van}{11.78} \hlcov{gaal}{7.05} \hlcov{was}{9.49} \hlcov{a}{0.10} \hlcov{man}{0.12} \hlcov{with}{0.06} \hlcov{a}{0.05} \hlcov{plan}{0.14} \hlcov{.}{0.72} \hlcov{the}{1.04} \hlcov{first}{0.24} \hlcov{season}{0.00} \hlcov{,}{0.00} \hlcov{he}{0.50} \hlcov{stated}{0.11} \hlcov{,}{0.00} \hlcov{would}{0.05} \hlcov{see}{0.02} \hlcov{him}{0.01} \hlcov{deliver}{0.05} \hlcov{manchester}{18.07} \hlcov{united}{23.84} \hlcov{back}{1.27} \hlcov{into}{0.12} \hlcov{their}{1.09} \hlcov{rightful}{0.19} \hlcov{place}{0.01} \hlcov{in}{0.37} \hlcov{the}{0.86} \hlcov{champions}{0.47} \hlcov{league}{1.94} \hlcov{.}{19.17} \hlcov{he}{4.96} \hlcov{would}{0.01} \hlcov{restore}{0.18} \hlcov{them}{0.04} \hlcov{to}{0.02} \hlcov{the}{0.09} \hlcov{premier}{0.18} \hlcov{league}{0.00} \hlcov{top}{0.14} \hlcov{four}{0.16} \hlcov{but}{0.10} \hlcov{\textit{loftier}}{2.92} \hlcov{aims}{0.02} \hlcov{of}{0.01} \hlcov{silverware}{0.06} \hlcov{would}{0.00} \hlcov{have}{0.01} \hlcov{to}{0.00} \hlcov{wait}{0.37} \hlcov{.}{0.36} \hlcov{his}{2.16} \hlcov{three-year}{0.54} \hlcov{vision}{0.15} \hlcov{would}{0.00} \hlcov{allow}{0.00} \hlcov{for}{0.00} \hlcov{such}{0.01} \hlcov{thoughts}{0.00} \hlcov{but}{0.10} \hlcov{,}{0.00} \hlcov{first}{0.16} \hlcov{things}{0.00} \hlcov{first}{0.01} \hlcov{,}{0.00} \hlcov{united}{7.71} \hlcov{needed}{26.57} \hlcov{to}{34.61} \hlcov{be}{34.71} \hlcov{dining}{35.31} \hlcov{from}{35.35} \hlcov{european}{34.97} \hlcov{football}{35.79} \hlcov{'s}{35.49} \hlcov{top}{33.40} \hlcov{table}{36.60} \hlcov{again}{35.31} \hlcov{.}{35.54} \hlcov{louis}{27.94} \hlcov{van}{29.27} \hlcov{gaal}{28.37} \hlcov{is}{25.29} \hlcov{close}{35.12} \hlcov{to}{35.44} \hlcov{delivering}{35.67} \hlcov{his}{32.92} \hlcov{\textit{first-season}}{20.34} \hlcov{aim}{38.54} \hlcov{of}{35.13} \hlcov{returning}{35.61} \hlcov{man}{35.03} \hlcov{united}{37.11} \hlcov{into}{34.63} \hlcov{champions}{33.35} \hlcov{league}{33.46} \hlcov{.}{16.45} \hlcov{wayne}{16.60} \hlcov{rooney}{0.58} \hlcov{smashes}{0.06} \hlcov{home}{0.03} \hlcov{during}{4.91} \hlcov{manchester}{7.82} \hlcov{united}{8.76} \hlcov{'s}{4.98} \hlcov{3-1}{35.45} \hlcov{win}{0.02} \hlcov{over}{12.12} \hlcov{aston}{40.00} \hlcov{villa}{26.26} \hlcov{on}{18.86} \hlcov{saturday}{35.44} \hlcov{.}{38.17} \hlcov{united}{15.11} \hlcov{'s}{29.13} \hlcov{win}{0.35} \hlcov{over}{0.02} \hlcov{aston}{1.86} \hlcov{villa}{9.23} \hlcov{took}{0.01} \hlcov{them}{0.00} \hlcov{third}{0.02} \hlcov{,}{0.01} \hlcov{eight}{0.26} \hlcov{points}{0.00} \hlcov{ahead}{0.00} \hlcov{of}{0.00} \hlcov{fifth-placed}{0.17} \hlcov{liverpool}{0.23} \hlcov{in}{0.04} \hlcov{the}{0.07} \hlcov{table}{0.16} \hlcov{.}{0.22} \hlcov{april}{1.82} \hlcov{12}{0.00} \hlcov{manchester}{0.11} \hlcov{city}{2.68} \hlcov{(}{1.85} \hlcov{h}{0.01} \hlcov{)}{0.00} \hlcov{.}{0.21} \hlcov{april}{0.84} \hlcov{18}{0.00} \hlcov{chelsea}{0.07} \hlcov{(}{0.57} \hlcov{a}{0.00} \hlcov{)}{0.00} \hlcov{.}{0.41} \hlcov{april}{0.81} \hlcov{26}{0.00} \hlcov{everton}{0.25} \hlcov{(}{0.12} \hlcov{a}{0.00} \hlcov{)}{0.00} \hlcov{.}{0.16} \hlcov{may}{0.42} \hlcov{2}{0.00} \hlcov{west}{0.31} \hlcov{bromwich}{0.00} \hlcov{albion}{0.00} \hlcov{(}{3.01} \hlcov{h}{0.04} \hlcov{)}{0.00} \hlcov{.}{0.07} \hlcov{may}{0.32} \hlcov{9}{0.01} \hlcov{crystal}{0.40} \hlcov{palace}{0.04} \hlcov{(}{0.01} \hlcov{a}{0.00} \hlcov{)}{0.00} \hlcov{.}{0.18} \hlcov{may}{0.25} \hlcov{17}{0.01} \hlcov{arsenal}{0.20} \hlcov{(}{10.93} \hlcov{h}{0.02} \hlcov{)}{0.00} \hlcov{.}{0.04} \hlcov{may}{0.30} \hlcov{24}{0.01} \hlcov{hull}{0.44} \hlcov{city}{0.01} \hlcov{(}{0.10} \hlcov{a}{0.03} \hlcov{)}{0.00} \hlcov{.}{0.21} \hlcov{one}{0.16} \hlcov{season}{0.01} \hlcov{out}{0.00} \hlcov{of}{0.00} \hlcov{the}{0.06} \hlcov{champions}{0.04} \hlcov{league}{0.06} \hlcov{was}{0.00} \hlcov{far}{0.00} \hlcov{from}{0.00} \hlcov{ideal}{0.04} \hlcov{,}{0.00} \hlcov{but}{0.00} \hlcov{two}{0.08} \hlcov{seasons}{0.00} \hlcov{would}{0.00} \hlcov{be}{0.03} \hlcov{an}{0.07} \hlcov{absolute}{0.00} \hlcov{disaster}{0.00} \hlcov{and}{0.00} \hlcov{something}{0.00} \hlcov{,}{0.00} \hlcov{he}{0.00} \hlcov{understood}{0.00} \hlcov{,}{0.00} \hlcov{that}{0.00} \hlcov{would}{0.00} \hlcov{not}{0.00} \hlcov{be}{0.00} \hlcov{tolerated}{0.03} \hlcov{.}{0.04} \hlcov{in}{0.29} \hlcov{november}{0.08} \hlcov{,}{0.00} \hlcov{even}{0.02} \hlcov{that}{0.00} \hlcov{was}{0.00} \hlcov{looking}{0.01} \hlcov{optimistic}{0.02} \hlcov{.}{0.01} \hlcov{a}{0.20} \hlcov{1-0}{0.13} \hlcov{defeat}{0.00} \hlcov{to}{0.00} \hlcov{manchester}{0.20} \hlcov{city}{0.87} \hlcov{meant}{0.00} \hlcov{that}{0.01} \hlcov{united}{0.35} \hlcov{had}{9.50} \hlcov{taken}{0.03} \hlcov{just}{0.03} \hlcov{13}{0.11} \hlcov{points}{0.00} \hlcov{from}{0.00} \hlcov{their}{0.20} \hlcov{opening}{0.00} \hlcov{10}{0.00} \hlcov{matches}{0.00} \hlcov{-}{0.00} \hlcov{it}{0.14} \hlcov{was}{0.01} \hlcov{their}{0.05} \hlcov{worst}{0.01} \hlcov{start}{0.00} \hlcov{to}{0.00} \hlcov{a}{0.03} \hlcov{league}{0.01} \hlcov{campaign}{0.01} \hlcov{since}{0.00} \hlcov{1986}{0.00} \hlcov{,}{0.00} \hlcov{when}{0.00} \hlcov{ron}{0.16} \hlcov{atkinson}{0.00} \hlcov{only}{0.00} \hlcov{lasted}{0.00} \hlcov{until}{0.00} \hlcov{november}{0.00} \hlcov{.}{0.00}
\newline \rule{\columnwidth}{0.4pt}
\textbf{Reference Summary:} \newline man united have an eight-point cushion from \textit{fifth-place} liverpool .\newline van gaal looks likely to deliver on his promise of top four finish .\newline but the dutchman has a three-year vision mapped out .\newline next season will have to see united mount sustained challenge for title .\newline they must also reach the later stages of the champions league .
\newline \rule{\columnwidth}{0.4pt}
\textbf{Baseline:} \newline
manchester united beat aston villa 3-1 at old trafford on saturday . \newline
louis van gaal is close to delivering his UNK aim of returning man united into the premier league top four . \newline
louis van gaal is close to delivering his UNK aim of returning man united into champions league .
\newline \rule{\columnwidth}{0.4pt}
\textbf{Pointer-Generator, No Coverage:} \newline
louis van gaal is close to delivering his \textit{first-season} aim of returning man united into champions league. \newline
united 's win over aston villa took them third , eight points ahead of fifth-placed liverpool in the table . \newline
louis van gaal is close to delivering his \textit{first-season} aim of returning man united into champions league.
\newline \rule{\columnwidth}{0.4pt}
\textbf{Pointer-Generator, With Coverage:} \newline  \hlpgen{manchester}{3.34} \hlpgen{united}{11.45} \hlpgen{beat}{36.26} \hlpgen{aston}{7.82} \hlpgen{villa}{1.86} \hlpgen{3-1}{8.01} \hlpgen{at}{38.87} \hlpgen{old}{8.19} \hlpgen{trafford}{1.94} \hlpgen{on}{36.33} \hlpgen{saturday}{10.52} \hlpgen{.}{39.36}\newline \hlpgen{louis}{9.46} \hlpgen{van}{4.85} \hlpgen{gaal}{3.32} \hlpgen{is}{32.81} \hlpgen{close}{13.66} \hlpgen{to}{32.44} \hlpgen{delivering}{23.35} \hlpgen{his}{11.64} \hlpgen{\textit{first-season}}{3.37} \hlpgen{aim}{3.85} \hlpgen{of}{20.89} \hlpgen{returning}{17.49} \hlpgen{man}{15.07} \hlpgen{united}{8.70} \hlpgen{into}{27.22} \hlpgen{champions}{3.12} \hlpgen{league}{3.95} \hlpgen{.}{33.62}\newline \hlpgen{united}{12.72} \hlpgen{needed}{34.29} \hlpgen{to}{18.08} \hlpgen{be}{13.29} \hlpgen{dining}{10.40} \hlpgen{from}{14.47} \hlpgen{european}{12.11} \hlpgen{football}{4.40} \hlpgen{'s}{4.44} \hlpgen{top}{0.90} \hlpgen{table}{1.92} \hlpgen{again}{7.82} \hlpgen{.}{33.93}\newline
\end{boxedminipage}
\caption{In this example, both our baseline model and final model produce a completely abstractive first sentence, using a novel word \textit{beat}.}
\label{fig_manu}
\end{figure*}

\begin{figure*}
\begin{boxedminipage}{\textwidth}
\textbf{Article (truncated):}  \hlcov{having}{0.19} \hlcov{been}{0.00} \hlcov{on}{3.26} \hlcov{the}{0.15} \hlcov{receiving}{0.07} \hlcov{end}{0.00} \hlcov{of}{0.00} \hlcov{a}{0.41} \hlcov{6-1}{1.67} \hlcov{thumping}{0.00} \hlcov{,}{0.00} \hlcov{a}{0.04} \hlcov{defeat}{0.28} \hlcov{like}{0.00} \hlcov{that}{0.01} \hlcov{could}{0.01} \hlcov{be}{0.00} \hlcov{justifiably}{0.15} \hlcov{met}{0.58} \hlcov{with}{0.00} \hlcov{a}{0.01} \hlcov{backlash}{0.01} \hlcov{by}{0.00} \hlcov{angry}{0.82} \hlcov{supporters}{0.58} \hlcov{.}{0.09} \hlcov{watching}{4.44} \hlcov{a}{0.14} \hlcov{3-1}{0.14} \hlcov{first}{0.00} \hlcov{leg}{0.00} \hlcov{aggregate}{0.01} \hlcov{advantage}{0.00} \hlcov{turn}{0.00} \hlcov{into}{0.00} \hlcov{a}{0.01} \hlcov{7-4}{0.03} \hlcov{deficit}{0.07} \hlcov{come}{0.00} \hlcov{the}{0.01} \hlcov{end}{0.08} \hlcov{of}{0.00} \hlcov{the}{0.06} \hlcov{reverse}{0.07} \hlcov{encounter}{0.02} \hlcov{too}{0.04} \hlcov{could}{0.01} \hlcov{send}{0.03} \hlcov{many}{0.14} \hlcov{fans}{0.99} \hlcov{\textit{apoplectic}}{1.82} \hlcov{at}{0.08} \hlcov{the}{0.15} \hlcov{capitulation}{0.06} \hlcov{of}{0.00} \hlcov{their}{0.72} \hlcov{side}{4.50} \hlcov{.}{0.11} \hlcov{however}{8.91} \hlcov{that}{30.82} \hlcov{does}{31.48} \hlcov{n't}{31.28} \hlcov{appear}{31.25} \hlcov{the}{31.13} \hlcov{case}{31.53} \hlcov{for}{31.29} \hlcov{those}{31.25} \hlcov{devoted}{31.54} \hlcov{to}{31.29} \hlcov{porto}{32.82} \hlcov{.}{31.06} \hlcov{porto}{21.65} \hlcov{supporters}{14.68} \hlcov{gave}{0.12} \hlcov{their}{0.03} \hlcov{team}{0.36} \hlcov{a}{0.09} \hlcov{hero}{0.09} \hlcov{'s}{0.13} \hlcov{welcome}{0.12} \hlcov{following}{1.76} \hlcov{their}{0.01} \hlcov{6-1}{0.13} \hlcov{defeat}{0.00} \hlcov{at}{0.00} \hlcov{bayern}{0.27} \hlcov{munich}{0.00} \hlcov{on}{0.01} \hlcov{tuesday}{0.04} \hlcov{.}{0.02} \hlcov{porto}{10.94} \hlcov{star}{13.41} \hlcov{striker}{19.63} \hlcov{jackson}{40.00} \hlcov{martinez}{31.39} \hlcov{was}{24.91} \hlcov{one}{25.03} \hlcov{of}{27.56} \hlcov{many}{25.99} \hlcov{players}{26.78} \hlcov{to}{31.86} \hlcov{look}{30.97} \hlcov{perplexed}{30.82} \hlcov{by}{31.26} \hlcov{their}{11.18} \hlcov{warm}{27.04} \hlcov{reception}{33.83} \hlcov{.}{31.16} \hlcov{porto}{13.34} \hlcov{boss}{18.74} \hlcov{\textit{julen}}{28.44} \hlcov{\textit{lopetegui}}{29.95} \hlcov{(}{0.37} \hlcov{left}{0.06} \hlcov{)}{0.00} \hlcov{was}{17.72} \hlcov{hugged}{36.91} \hlcov{by}{30.90} \hlcov{fans}{26.06} \hlcov{congratulating}{21.87} \hlcov{him}{31.14} \hlcov{on}{31.26} \hlcov{their}{25.30} \hlcov{champions}{19.21} \hlcov{league}{31.29} \hlcov{run}{31.12} \hlcov{.}{28.92} \hlcov{police}{8.35} \hlcov{escorts}{27.14} \hlcov{were}{33.22} \hlcov{needed}{33.41} \hlcov{to}{31.30} \hlcov{keep}{31.35} \hlcov{the}{30.78} \hlcov{delirious}{26.69} \hlcov{supporters}{32.47} \hlcov{at}{30.32} \hlcov{bay}{31.60} \hlcov{as}{31.22} \hlcov{the}{0.40} \hlcov{porto}{4.45} \hlcov{team}{0.53} \hlcov{bus}{0.03} \hlcov{drove}{0.01} \hlcov{past}{0.05} \hlcov{.}{0.01} \hlcov{the}{3.23} \hlcov{team}{3.50} \hlcov{bus}{0.14} \hlcov{was}{0.11} \hlcov{met}{3.99} \hlcov{with}{0.00} \hlcov{a}{0.01} \hlcov{cacophony}{0.01} \hlcov{of}{0.00} \hlcov{noise}{0.29} \hlcov{from}{0.05} \hlcov{porto}{4.91} \hlcov{supporters}{2.44} \hlcov{proudly}{1.01} \hlcov{chanting}{0.94} \hlcov{about}{0.04} \hlcov{their}{0.77} \hlcov{club}{3.97} \hlcov{.}{0.08} \hlcov{on}{6.97} \hlcov{their}{0.37} \hlcov{return}{0.79} \hlcov{from}{0.04} \hlcov{a}{0.01} \hlcov{humiliating}{0.01} \hlcov{champions}{0.09} \hlcov{league}{0.05} \hlcov{quarter-final}{0.22} \hlcov{loss}{0.00} \hlcov{at}{0.00} \hlcov{the}{0.02} \hlcov{hands}{0.00} \hlcov{of}{0.00} \hlcov{bayern}{3.18} \hlcov{munich}{0.00} \hlcov{on}{0.03} \hlcov{tuesday}{0.09} \hlcov{night}{0.00} \hlcov{,}{0.04} \hlcov{the}{9.02} \hlcov{squad}{2.33} \hlcov{were}{0.03} \hlcov{given}{0.28} \hlcov{a}{0.01} \hlcov{heroes}{0.03} \hlcov{reception}{0.05} \hlcov{as}{0.03} \hlcov{they}{0.04} \hlcov{arrived}{0.04} \hlcov{back}{0.00} \hlcov{in}{0.00} \hlcov{portugal}{0.02} \hlcov{.}{0.09} \hlcov{in}{1.93} \hlcov{the}{0.00} \hlcov{early}{0.01} \hlcov{hours}{0.00} \hlcov{of}{0.00} \hlcov{wednesday}{0.04} \hlcov{morning}{0.00} \hlcov{,}{0.02} \hlcov{fans}{6.39} \hlcov{mobbed}{0.59} \hlcov{the}{0.06} \hlcov{squad}{0.12} \hlcov{congratulating}{0.03} \hlcov{them}{0.19} \hlcov{on}{0.04} \hlcov{their}{0.44} \hlcov{run}{10.77} \hlcov{in}{0.54} \hlcov{the}{0.02} \hlcov{tournament}{0.05} \hlcov{.}{0.05} \hlcov{star}{4.75} \hlcov{striker}{1.32} \hlcov{jackson}{2.37} \hlcov{martinez}{1.48} \hlcov{and}{0.63} \hlcov{ricardo}{0.13} \hlcov{\textit{quaresma}}{0.04} \hlcov{were}{0.12} \hlcov{one}{2.05} \hlcov{of}{3.72} \hlcov{many}{2.91} \hlcov{porto}{7.25} \hlcov{players}{1.02} \hlcov{who}{0.55} \hlcov{looked}{0.02} \hlcov{perplexed}{0.48} \hlcov{as}{0.05} \hlcov{they}{0.01} \hlcov{were}{0.00} \hlcov{hugged}{0.04} \hlcov{by}{0.33} \hlcov{fans}{11.83} \hlcov{before}{7.46} \hlcov{they}{0.00} \hlcov{making}{0.00} \hlcov{their}{0.00} \hlcov{way}{0.01} \hlcov{on}{0.00} \hlcov{to}{1.97} \hlcov{the}{0.11} \hlcov{team}{0.23} \hlcov{bus}{0.02} \hlcov{-}{0.00} \hlcov{set}{0.06} \hlcov{upon}{0.00} \hlcov{a}{0.01} \hlcov{cacophony}{0.01} \hlcov{of}{0.00} \hlcov{\textit{fiercely-proud}}{0.06} \hlcov{chanting}{0.05} \hlcov{.}{0.01} \hlcov{it}{12.43} \hlcov{was}{0.01} \hlcov{the}{0.28} \hlcov{first}{0.02} \hlcov{time}{0.00} \hlcov{that}{0.00} \hlcov{porto}{4.14} \hlcov{,}{0.52} \hlcov{who}{0.00} \hlcov{had}{0.00} \hlcov{been}{0.00} \hlcov{unbeaten}{0.11} \hlcov{in}{0.00} \hlcov{this}{0.00} \hlcov{season}{0.00} \hlcov{'s}{0.00} \hlcov{tournament}{0.01} \hlcov{up}{0.00} \hlcov{until}{0.00} \hlcov{tuesday}{0.00} \hlcov{night}{0.00} \hlcov{,}{0.00} \hlcov{had}{0.15} \hlcov{reached}{0.04} \hlcov{the}{0.00} \hlcov{quarter-finals}{0.00} \hlcov{of}{0.00} \hlcov{the}{0.00} \hlcov{champions}{0.01} \hlcov{league}{0.01} \hlcov{since}{0.00} \hlcov{the}{0.00} \hlcov{2008-09}{0.00} \hlcov{season}{0.00} \hlcov{.}{0.00}
\newline \rule{\columnwidth}{0.4pt}
\textbf{Reference Summary:} \newline bayern munich beat porto 6-1 in their champions league tie on tuesday .\newline result saw bayern win quarter-final encounter 7-4 on aggregate .\newline it was the first-time porto had reached that stage since the 2008-09 season .
\newline \rule{\columnwidth}{0.4pt}
\textbf{Baseline:} \newline
porto \error{beat} bayern munich \error{2-0} in the champions league on tuesday night . \newline
porto star \error{james} UNK was one of many players involved in the match . \newline
the squad were given a \error{trophy} as they arrived back in portugal .
\newline \rule{\columnwidth}{0.4pt}
\textbf{Pointer-Generator, No Coverage:} \newline
porto star striker jackson martinez was one of many players to look perplexed by their warm reception. \newline
porto star striker jackson martinez was one of many players to look perplexed by their warm reception. \newline
porto star striker jackson martinez was one of many players to look perplexed by their warm reception.
\newline \rule{\columnwidth}{0.4pt}
\textbf{Pointer-Generator, With Coverage:} \newline  \hlpgen{porto}{2.96} \hlpgen{star}{16.84} \hlpgen{striker}{6.74} \hlpgen{jackson}{2.68} \hlpgen{martinez}{1.84} \hlpgen{was}{17.83} \hlpgen{one}{15.28} \hlpgen{of}{14.29} \hlpgen{many}{8.98} \hlpgen{players}{10.26} \hlpgen{to}{27.60} \hlpgen{look}{16.92} \hlpgen{perplexed}{10.79} \hlpgen{by}{14.89} \hlpgen{their}{3.03} \hlpgen{warm}{4.68} \hlpgen{reception}{3.91} \hlpgen{.}{31.07}\newline \hlpgen{porto}{7.11} \hlpgen{boss}{11.85} \hlpgen{julen}{3.80} \hlpgen{lopetegui}{0.30} \hlpgen{was}{10.97} \hlpgen{hugged}{10.86} \hlpgen{by}{17.35} \hlpgen{fans}{3.38} \hlpgen{congratulating}{11.97} \hlpgen{him}{1.34} \hlpgen{on}{8.47} \hlpgen{their}{4.51} \hlpgen{champions}{2.35} \hlpgen{league}{2.10} \hlpgen{run}{6.62} \hlpgen{.}{26.98}\newline \hlpgen{however}{12.62} \hlpgen{that}{13.36} \hlpgen{does}{18.37} \hlpgen{n't}{13.24} \hlpgen{appear}{9.41} \hlpgen{the}{3.34} \hlpgen{case}{1.67} \hlpgen{for}{2.55} \hlpgen{those}{1.53} \hlpgen{devoted}{3.08} \hlpgen{to}{3.11} \hlpgen{porto}{0.78} \hlpgen{.}{18.15}\newline \hlpgen{police}{21.02} \hlpgen{escorts}{8.57} \hlpgen{were}{13.41} \hlpgen{needed}{3.38} \hlpgen{to}{13.99} \hlpgen{keep}{4.21} \hlpgen{the}{4.01} \hlpgen{delirious}{3.29} \hlpgen{supporters}{1.88} \hlpgen{at}{3.78} \hlpgen{bay}{0.55} \hlpgen{.}{27.40}\newline
\end{boxedminipage}
\caption{The baseline model makes several factual inaccuracies: it claims \textit{porto} beat \textit{bayern munich} not vice versa, the score is changed from \textit{7-4} to \textit{2-0}, \textit{jackson} is changed to \textit{james} and \textit{a heroes reception} is replaced with \textit{a trophy}.
Our final model produces sentences that are individually accurate, but they do not make sense as a whole.
Note that the final model omits the parenthesized phrase \textit{( left )} from its second sentence.}
\label{fig_porto}
\end{figure*}

\begin{figure*}
\begin{boxedminipage}{\textwidth}
\textbf{Article:}  \hlcov{(}{0.00} \hlcov{cnn}{0.01} \hlcov{)}{0.00} \hlcov{''}{0.01} \hlcov{it}{0.35} \hlcov{'s}{0.04} \hlcov{showtime}{0.35} \hlcov{!}{0.01} \hlcov{''}{0.01} \hlcov{michael}{11.36} \hlcov{keaton}{28.68} \hlcov{paid}{16.47} \hlcov{homage}{27.82} \hlcov{--}{26.13} \hlcov{ever}{0.40} \hlcov{so}{0.08} \hlcov{slightly}{0.03} \hlcov{--}{0.03} \hlcov{to}{1.34} \hlcov{his}{21.09} \hlcov{roles}{22.72} \hlcov{in}{29.20} \hlcov{``}{24.90} \hlcov{\textit{beetlejuice}}{40.00} \hlcov{''}{28.56} \hlcov{and}{28.19} \hlcov{``}{30.78} \hlcov{batman}{14.40} \hlcov{''}{28.45} \hlcov{in}{27.85} \hlcov{his}{28.11} \hlcov{third}{29.82} \hlcov{turn}{30.77} \hlcov{hosting}{28.74} \hlcov{``}{24.63} \hlcov{saturday}{35.64} \hlcov{night}{28.60} \hlcov{live}{28.65} \hlcov{''}{28.64} \hlcov{this}{24.68} \hlcov{weekend}{0.52} \hlcov{.}{3.44} \hlcov{keaton}{22.11} \hlcov{acknowledged}{27.54} \hlcov{in}{28.54} \hlcov{his}{28.83} \hlcov{opening}{28.85} \hlcov{monologue}{29.19} \hlcov{that}{27.78} \hlcov{a}{27.94} \hlcov{lot}{28.02} \hlcov{has}{28.61} \hlcov{changed}{28.33} \hlcov{since}{29.06} \hlcov{he}{30.04} \hlcov{first}{23.39} \hlcov{hosted}{33.75} \hlcov{the}{29.57} \hlcov{comedy}{28.02} \hlcov{sketch}{26.90} \hlcov{show}{26.59} \hlcov{in}{34.41} \hlcov{1982}{29.94} \hlcov{.}{28.88} \hlcov{``}{2.45} \hlcov{i}{0.01} \hlcov{had}{0.00} \hlcov{a}{0.04} \hlcov{baby}{0.91} \hlcov{--}{0.00} \hlcov{he}{0.06} \hlcov{'s}{0.09} \hlcov{31}{0.09} \hlcov{.}{0.00} \hlcov{i}{0.15} \hlcov{also}{0.01} \hlcov{have}{0.00} \hlcov{a}{0.02} \hlcov{new}{0.09} \hlcov{girlfriend}{0.07} \hlcov{--}{0.00} \hlcov{she}{0.07} \hlcov{'s}{0.01} \hlcov{28}{0.08} \hlcov{,}{0.00} \hlcov{''}{0.00} \hlcov{he}{0.16} \hlcov{said}{0.02} \hlcov{.}{0.73} \hlcov{fans}{4.81} \hlcov{who}{0.00} \hlcov{were}{0.00} \hlcov{hoping}{0.00} \hlcov{for}{0.01} \hlcov{a}{0.05} \hlcov{full-blown}{0.04} \hlcov{revival}{0.07} \hlcov{of}{0.01} \hlcov{keaton}{0.46} \hlcov{'s}{1.85} \hlcov{most}{0.10} \hlcov{memorable}{0.14} \hlcov{characters}{0.16} \hlcov{might}{0.00} \hlcov{have}{0.00} \hlcov{been}{0.27} \hlcov{a}{0.00} \hlcov{little}{0.00} \hlcov{disappointed}{0.00} \hlcov{.}{0.01} \hlcov{snl}{4.86} \hlcov{cast}{0.10} \hlcov{members}{0.07} \hlcov{\textit{taran}}{0.03} \hlcov{\textit{killam}}{0.01} \hlcov{and}{0.01} \hlcov{bobby}{0.08} \hlcov{moynihan}{0.00} \hlcov{begged}{0.01} \hlcov{the}{0.08} \hlcov{actor}{0.16} \hlcov{with}{0.02} \hlcov{a}{0.03} \hlcov{song}{0.17} \hlcov{to}{0.08} \hlcov{``}{2.16} \hlcov{play}{1.80} \hlcov{''}{0.00} \hlcov{batman}{0.28} \hlcov{and}{0.14} \hlcov{\textit{beetlejuice}}{0.03} \hlcov{with}{0.15} \hlcov{them}{0.00} \hlcov{.}{0.10} \hlcov{all}{1.19} \hlcov{they}{0.02} \hlcov{got}{0.00} \hlcov{in}{0.00} \hlcov{response}{0.03} \hlcov{were}{0.00} \hlcov{a}{0.00} \hlcov{couple}{0.02} \hlcov{of}{0.00} \hlcov{one-liners}{0.02} \hlcov{.}{0.00} \hlcov{overall}{1.09} \hlcov{,}{0.00} \hlcov{keaton}{7.13} \hlcov{'s}{10.76} \hlcov{performance}{0.08} \hlcov{drew}{0.08} \hlcov{high}{0.00} \hlcov{marks}{0.00} \hlcov{from}{0.00} \hlcov{viewers}{0.10} \hlcov{and}{0.00} \hlcov{critics}{0.81} \hlcov{for}{0.00} \hlcov{its}{0.00} \hlcov{``}{0.05} \hlcov{deadpan}{0.01} \hlcov{''}{0.00} \hlcov{manner}{0.00} \hlcov{and}{0.00} \hlcov{``}{0.05} \hlcov{unpredictable}{0.19} \hlcov{\textit{weirdness}}{0.03} \hlcov{,}{0.00} \hlcov{''}{0.00} \hlcov{in}{0.03} \hlcov{the}{0.10} \hlcov{words}{0.01} \hlcov{of}{0.00} \hlcov{\textit{a.v}}{0.09} \hlcov{.}{0.03} \hlcov{club}{3.23} \hlcov{'s}{0.00} \hlcov{dennis}{0.05} \hlcov{perkins}{0.00} \hlcov{.}{0.18} \hlcov{fans}{2.54} \hlcov{also}{0.00} \hlcov{delighted}{0.00} \hlcov{in}{0.00} \hlcov{a}{0.10} \hlcov{cameo}{0.01} \hlcov{from}{0.03} \hlcov{``}{0.59} \hlcov{walking}{0.52} \hlcov{dead}{0.01} \hlcov{''}{0.00} \hlcov{star}{0.03} \hlcov{norman}{0.14} \hlcov{\textit{reedus}}{0.01} \hlcov{during}{0.05} \hlcov{weekend}{0.04} \hlcov{update}{0.02} \hlcov{.}{0.13} \hlcov{keaton}{8.27} \hlcov{scored}{0.27} \hlcov{some}{0.00} \hlcov{laughs}{0.00} \hlcov{from}{0.00} \hlcov{the}{0.01} \hlcov{audience}{0.05} \hlcov{as}{0.01} \hlcov{an}{0.00} \hlcov{ad}{0.04} \hlcov{executive}{0.00} \hlcov{who}{0.00} \hlcov{'s}{0.00} \hlcov{not}{0.03} \hlcov{very}{0.00} \hlcov{good}{0.00} \hlcov{at}{0.00} \hlcov{his}{0.09} \hlcov{job}{0.57} \hlcov{,}{0.02} \hlcov{a}{0.01} \hlcov{confused}{0.03} \hlcov{grandfather}{0.10} \hlcov{and}{0.00} \hlcov{a}{0.00} \hlcov{high}{0.02} \hlcov{school}{0.01} \hlcov{teacher}{0.00} \hlcov{who}{0.00} \hlcov{gets}{0.00} \hlcov{asked}{0.00} \hlcov{to}{0.02} \hlcov{the}{0.16} \hlcov{prom}{0.90} \hlcov{in}{0.01} \hlcov{a}{0.08} \hlcov{riff}{0.01} \hlcov{on}{0.02} \hlcov{the}{0.01} \hlcov{romantic}{0.05} \hlcov{comedy}{0.04} \hlcov{``}{0.06} \hlcov{she}{0.08} \hlcov{'s}{0.00} \hlcov{all}{0.01} \hlcov{that}{0.00} \hlcov{.}{0.00} \hlcov{''}{0.00} \hlcov{other}{1.70} \hlcov{\textit{crowd-pleasing}}{0.11} \hlcov{spots}{0.01} \hlcov{included}{0.00} \hlcov{a}{0.01} \hlcov{scientology}{0.13} \hlcov{parody}{0.02} \hlcov{music}{0.01} \hlcov{video}{0.00} \hlcov{and}{0.01} \hlcov{a}{0.00} \hlcov{news}{0.00} \hlcov{conference}{0.00} \hlcov{\textit{spoofing}}{0.00} \hlcov{the}{0.02} \hlcov{ncaa}{0.13} \hlcov{\textit{student-athlete}}{0.00} \hlcov{debate}{0.03} \hlcov{.}{0.01} \hlcov{the}{2.95} \hlcov{show}{0.08} \hlcov{also}{0.01} \hlcov{poked}{0.00} \hlcov{fun}{0.00} \hlcov{at}{0.00} \hlcov{cnn}{0.30} \hlcov{with}{0.00} \hlcov{cast}{0.03} \hlcov{member}{0.00} \hlcov{cecily}{0.02} \hlcov{strong}{0.01} \hlcov{playing}{0.03} \hlcov{anchor}{0.05} \hlcov{brooke}{0.01} \hlcov{baldwin}{0.00} \hlcov{.}{0.00} \hlcov{baldwin}{1.50} \hlcov{said}{0.00} \hlcov{on}{0.00} \hlcov{twitter}{0.05} \hlcov{that}{0.00} \hlcov{she}{0.18} \hlcov{'s}{0.00} \hlcov{taking}{0.00} \hlcov{it}{0.00} \hlcov{as}{0.02} \hlcov{a}{0.00} \hlcov{``}{0.02} \hlcov{crazy}{0.01} \hlcov{compliment}{0.00} \hlcov{''}{0.00} \hlcov{and}{0.02} \hlcov{shared}{0.01} \hlcov{a}{0.01} \hlcov{clip}{0.01} \hlcov{from}{0.12} \hlcov{the}{0.03} \hlcov{skit}{0.04} \hlcov{.}{0.00}
\newline \rule{\columnwidth}{0.4pt}
\textbf{Reference Summary:} \newline michael keaton hosted `` saturday night live '' for the first time in 1982 .\newline in 2015 , his nods to starring roles in `` \textit{beetlejuice} '' and `` batman '' are brief .
\newline \rule{\columnwidth}{0.4pt}
\textbf{Baseline:} \newline
`` snl '' wins `` saturday night live '' . \newline
keaton acknowledged in his opening `` saturday night live '' . \newline
keaton acknowledged in his opening night 's `` UNK '' . \newline
the show also pokes fun at cnn with cast member \error{mariah carey} .
\newline \rule{\columnwidth}{0.4pt}
\textbf{Pointer-Generator, No Coverage:} \newline
`` \textit{beetlejuice} '' michael keaton paid homage to his roles in `` \textit{beetlejuice} '' and `` batman '' in his third turn hosting `` saturday night live '' this weekend . \newline
keaton acknowledged in his opening monologue that a lot has changed since he first hosted the comedy sketch show in 1982 .
\newline \rule{\columnwidth}{0.4pt}
\textbf{Pointer-Generator, With Coverage:} \newline  \hlpgen{michael}{8.40} \hlpgen{keaton}{1.78} \hlpgen{paid}{28.72} \hlpgen{homage}{10.49} \hlpgen{to}{29.77} \hlpgen{his}{10.51} \hlpgen{roles}{5.59} \hlpgen{in}{18.27} \hlpgen{``}{3.76} \hlpgen{\textit{beetlejuice}}{0.51} \hlpgen{''}{6.97} \hlpgen{and}{6.76} \hlpgen{``}{4.64} \hlpgen{batman}{0.40} \hlpgen{''}{29.84} \hlpgen{in}{20.23} \hlpgen{his}{3.42} \hlpgen{third}{1.16} \hlpgen{turn}{8.86} \hlpgen{hosting}{12.80} \hlpgen{``}{17.68} \hlpgen{saturday}{0.96} \hlpgen{night}{0.15} \hlpgen{live}{0.69} \hlpgen{''}{25.60} \hlpgen{.}{26.63}\newline \hlpgen{keaton}{9.96} \hlpgen{acknowledged}{23.20} \hlpgen{in}{9.44} \hlpgen{his}{1.96} \hlpgen{opening}{0.95} \hlpgen{monologue}{1.39} \hlpgen{that}{8.14} \hlpgen{a}{2.24} \hlpgen{lot}{0.25} \hlpgen{has}{1.55} \hlpgen{changed}{0.86} \hlpgen{since}{7.25} \hlpgen{he}{3.50} \hlpgen{first}{1.15} \hlpgen{hosted}{6.02} \hlpgen{the}{0.95} \hlpgen{comedy}{0.39} \hlpgen{sketch}{0.19} \hlpgen{show}{0.39} \hlpgen{in}{3.08} \hlpgen{1982}{0.25} \hlpgen{.}{14.89}\newline
\end{boxedminipage}
\caption{Baseline model replaces \textit{cecily strong} with \textit{mariah carey}, and produces generally nonsensical output.
The baseline model may be struggling with the out-of-vocabulary word \textit{beetlejuice}, or perhaps the unusual non-news format of the article.
Note that the final model omits \textit{-- ever so slightly --} from its first sentence.}
\label{fig_snl}
\end{figure*}

\begin{figure*}
\begin{boxedminipage}{\textwidth}
\textbf{Article (truncated):}  \hlcov{they}{1.75} \hlcov{are}{0.85} \hlcov{supposed}{0.03} \hlcov{to}{0.00} \hlcov{be}{0.03} \hlcov{the}{0.60} \hlcov{dream}{0.42} \hlcov{team}{0.24} \hlcov{who}{0.00} \hlcov{can}{0.04} \hlcov{solve}{0.14} \hlcov{the}{0.03} \hlcov{conundrum}{0.05} \hlcov{of}{0.00} \hlcov{how}{0.01} \hlcov{to}{0.01} \hlcov{win}{0.27} \hlcov{the}{0.00} \hlcov{election}{0.09} \hlcov{.}{0.00} \hlcov{but}{11.94} \hlcov{david}{14.65} \hlcov{cameron}{1.44} \hlcov{and}{0.19} \hlcov{boris}{4.06} \hlcov{johnson}{0.14} \hlcov{were}{0.06} \hlcov{left}{0.06} \hlcov{scratching}{0.00} \hlcov{their}{0.01} \hlcov{heads}{0.02} \hlcov{today}{0.54} \hlcov{as}{0.01} \hlcov{they}{0.85} \hlcov{struggled}{0.73} \hlcov{with}{0.00} \hlcov{a}{0.25} \hlcov{children}{1.56} \hlcov{'s}{0.00} \hlcov{jigsaw}{0.57} \hlcov{teaching}{0.21} \hlcov{toddlers}{0.12} \hlcov{about}{0.05} \hlcov{the}{0.01} \hlcov{seasons}{0.02} \hlcov{.}{0.39} \hlcov{as}{1.79} \hlcov{the}{1.48} \hlcov{london}{15.22} \hlcov{mayor}{35.35} \hlcov{tried}{32.94} \hlcov{to}{32.72} \hlcov{hammer}{33.92} \hlcov{ill-fitting}{30.78} \hlcov{pieces}{37.06} \hlcov{together}{34.20} \hlcov{with}{33.96} \hlcov{his}{26.74} \hlcov{hands}{35.98} \hlcov{,}{33.29} \hlcov{the}{4.50} \hlcov{prime}{16.81} \hlcov{minister}{34.46} \hlcov{tried}{33.32} \hlcov{out}{33.74} \hlcov{what}{34.28} \hlcov{could}{34.33} \hlcov{be}{33.94} \hlcov{a}{31.12} \hlcov{new}{33.58} \hlcov{election}{33.46} \hlcov{slogan}{40.00} \hlcov{,}{34.36} \hlcov{telling}{0.16} \hlcov{him}{0.00} \hlcov{:}{0.00} \hlcov{'}{0.04} \hlcov{if}{0.03} \hlcov{in}{0.00} \hlcov{doubt}{0.00} \hlcov{,}{0.00} \hlcov{wedge}{0.12} \hlcov{it}{0.01} \hlcov{in}{0.00} \hlcov{.}{0.00} \hlcov{'}{0.05} \hlcov{after}{0.95} \hlcov{being}{0.31} \hlcov{put}{0.47} \hlcov{right}{0.01} \hlcov{by}{0.02} \hlcov{a}{0.35} \hlcov{four-year-old}{2.07} \hlcov{who}{0.01} \hlcov{spotted}{0.08} \hlcov{their}{0.03} \hlcov{errors}{0.05} \hlcov{,}{0.02} \hlcov{the}{7.10} \hlcov{pair}{9.98} \hlcov{had}{0.07} \hlcov{more}{0.04} \hlcov{fun}{0.04} \hlcov{finger}{0.03} \hlcov{painting}{0.09} \hlcov{with}{0.15} \hlcov{tory}{1.86} \hlcov{blue}{0.18} \hlcov{paint}{0.26} \hlcov{.}{0.19} \hlcov{david}{6.65} \hlcov{cameron}{0.32} \hlcov{and}{1.18} \hlcov{boris}{3.67} \hlcov{johnson}{0.08} \hlcov{were}{0.17} \hlcov{left}{0.27} \hlcov{stumped}{0.06} \hlcov{by}{0.00} \hlcov{the}{0.10} \hlcov{puzzle}{0.77} \hlcov{at}{0.04} \hlcov{advantage}{0.18} \hlcov{children}{0.34} \hlcov{'s}{0.00} \hlcov{day}{0.01} \hlcov{nursery}{0.19} \hlcov{in}{0.04} \hlcov{\textit{surbiton}}{0.08} \hlcov{,}{0.00} \hlcov{as}{0.11} \hlcov{three-year-old}{3.03} \hlcov{stephanie}{1.05} \hlcov{looked}{0.01} \hlcov{on}{0.00} \hlcov{.}{0.00} \hlcov{when}{1.15} \hlcov{they}{0.90} \hlcov{tried}{1.33} \hlcov{to}{2.00} \hlcov{put}{0.03} \hlcov{the}{0.04} \hlcov{puzzle}{0.19} \hlcov{back}{0.01} \hlcov{together}{0.01} \hlcov{,}{0.01} \hlcov{they}{8.12} \hlcov{hit}{24.56} \hlcov{trouble}{28.39} \hlcov{after}{30.87} \hlcov{it}{27.95} \hlcov{proved}{25.91} \hlcov{to}{25.89} \hlcov{be}{28.33} \hlcov{more}{25.78} \hlcov{difficult}{27.20} \hlcov{than}{23.43} \hlcov{expected}{22.50} \hlcov{.}{28.13} \hlcov{the}{1.49} \hlcov{conservative}{6.98} \hlcov{duo}{0.77} \hlcov{made}{0.35} \hlcov{their}{0.00} \hlcov{first}{0.02} \hlcov{appearance}{0.01} \hlcov{together}{0.00} \hlcov{on}{0.29} \hlcov{the}{0.01} \hlcov{campaign}{0.22} \hlcov{trail}{0.00} \hlcov{with}{0.00} \hlcov{a}{0.01} \hlcov{visit}{0.22} \hlcov{to}{0.00} \hlcov{advantage}{0.01} \hlcov{day}{0.00} \hlcov{nursery}{0.29} \hlcov{in}{0.00} \hlcov{\textit{surbiton}}{0.01} \hlcov{,}{0.00} \hlcov{south}{0.03} \hlcov{west}{0.00} \hlcov{london}{0.00} \hlcov{.}{0.00} \hlcov{they}{5.84} \hlcov{were}{3.00} \hlcov{supposed}{0.05} \hlcov{to}{0.01} \hlcov{be}{0.03} \hlcov{highlighting}{0.10} \hlcov{tory}{0.78} \hlcov{plans}{0.08} \hlcov{to}{0.03} \hlcov{double}{0.02} \hlcov{free}{0.01} \hlcov{childcare}{0.02} \hlcov{for}{0.00} \hlcov{600,000}{0.05} \hlcov{children}{0.01} \hlcov{aged}{0.00} \hlcov{three}{0.00} \hlcov{and}{0.00} \hlcov{four}{0.00} \hlcov{.}{0.00} \hlcov{they}{4.20} \hlcov{sat}{0.76} \hlcov{at}{0.00} \hlcov{a}{0.00} \hlcov{table}{0.00} \hlcov{with}{0.00} \hlcov{joshua}{0.10} \hlcov{and}{0.00} \hlcov{stephanie}{0.02} \hlcov{,}{0.00} \hlcov{both}{0.06} \hlcov{aged}{0.00} \hlcov{three}{0.00} \hlcov{,}{0.00} \hlcov{and}{0.00} \hlcov{four-year-olds}{0.01} \hlcov{hamish}{0.00} \hlcov{and}{0.00} \hlcov{leo}{0.01} \hlcov{where}{0.00} \hlcov{they}{0.11} \hlcov{broke}{0.06} \hlcov{up}{0.01} \hlcov{a}{0.01} \hlcov{puzzle}{0.01} \hlcov{which}{0.00} \hlcov{depicted}{0.04} \hlcov{the}{0.00} \hlcov{four}{0.01} \hlcov{seasons}{0.00} \hlcov{.}{0.02} \hlcov{when}{0.55} \hlcov{they}{0.10} \hlcov{tried}{0.27} \hlcov{to}{0.24} \hlcov{put}{0.01} \hlcov{it}{0.01} \hlcov{back}{0.00} \hlcov{together}{0.00} \hlcov{,}{0.01} \hlcov{they}{0.96} \hlcov{hit}{2.89} \hlcov{trouble}{5.97} \hlcov{after}{3.50} \hlcov{it}{6.33} \hlcov{proved}{8.47} \hlcov{to}{8.08} \hlcov{be}{6.01} \hlcov{more}{7.82} \hlcov{difficult}{7.69} \hlcov{than}{11.57} \hlcov{expected}{11.90} \hlcov{.}{6.25}
\newline \rule{\columnwidth}{0.4pt}
\textbf{Reference Summary:} \newline david cameron and boris johnson visit nursery to promote childcare .\newline duo left \textit{flummoxed} by jigsaw aimed at teaching children about seasons .\newline london mayor resorted to brute force to hammer wrong pieces together .\newline prime minister gave his approval , telling boris : ' if in doubt , wedge it in ' .
\newline \rule{\columnwidth}{0.4pt}
\textbf{Baseline:} \newline
david cameron and boris johnson were left mystified by the puzzle . \newline
they were at the nursery in UNK to promote a tory pledge to double free childcare for \error{25} children aged three and four . \newline
the pair were at the nursery in UNK to promote a tory pledge to double free childcare for children aged three and four .
\newline \rule{\columnwidth}{0.4pt}
\textbf{Pointer-Generator, No Coverage:} \newline
london mayor tried to hammer ill-fitting pieces together with his hands . \newline
they were supposed to be highlighting tory plans to double free childcare . \newline
they were supposed to be highlighting tory plans to double free childcare . \newline
they were supposed to be highlighting tory plans to double free childcare .
\newline \rule{\columnwidth}{0.4pt}
\textbf{Pointer-Generator, With Coverage:} \newline  \hlpgen{london}{9.27} \hlpgen{mayor}{6.80} \hlpgen{tried}{31.69} \hlpgen{to}{13.57} \hlpgen{hammer}{6.71} \hlpgen{ill-fitting}{8.38} \hlpgen{pieces}{5.46} \hlpgen{together}{4.77} \hlpgen{with}{7.95} \hlpgen{his}{2.16} \hlpgen{hands}{1.10} \hlpgen{.}{35.26}\newline \hlpgen{prime}{13.02} \hlpgen{minister}{1.54} \hlpgen{tried}{20.89} \hlpgen{out}{7.26} \hlpgen{what}{6.36} \hlpgen{could}{4.42} \hlpgen{be}{4.05} \hlpgen{a}{2.89} \hlpgen{new}{1.06} \hlpgen{election}{0.75} \hlpgen{slogan}{2.53} \hlpgen{.}{36.67}\newline \hlpgen{but}{10.96} \hlpgen{they}{17.76} \hlpgen{hit}{25.45} \hlpgen{trouble}{4.14} \hlpgen{after}{24.77} \hlpgen{it}{8.67} \hlpgen{proved}{12.35} \hlpgen{to}{7.41} \hlpgen{be}{4.35} \hlpgen{more}{3.68} \hlpgen{difficult}{1.54} \hlpgen{than}{8.08} \hlpgen{expected}{3.38} \hlpgen{.}{36.41}\newline
\end{boxedminipage}
\caption{The baseline model appropriately replaces \textit{stumped} with novel word \textit{mystified}.
However, the reference summary chooses \textit{flummoxed} (also novel) so the choice of \textit{mystified} is not rewarded by the ROUGE metric.
The baseline model also incorrectly substitutes \textit{600,000} for \textit{25}.
In the final model's output we observe that the generation probability is largest at the beginning of sentences (especially the first verb) and on periods.
}
\label{fig_puzzle}
\end{figure*}

\begin{figure*}
\begin{boxedminipage}{\textwidth}
\textbf{Article (truncated):}  \hlcov{lagos}{0.11} \hlcov{,}{0.00} \hlcov{nigeria}{1.27} \hlcov{(}{0.00} \hlcov{cnn}{0.12} \hlcov{)}{0.00} \hlcov{a}{0.00} \hlcov{day}{0.26} \hlcov{after}{0.01} \hlcov{winning}{2.07} \hlcov{nigeria}{4.79} \hlcov{'s}{0.07} \hlcov{presidency}{0.05} \hlcov{,}{0.01} \hlcov{\textit{muhammadu}}{17.98} \hlcov{\textit{buhari}}{32.37} \hlcov{told}{24.20} \hlcov{cnn}{0.71} \hlcov{'s}{0.00} \hlcov{christiane}{0.45} \hlcov{amanpour}{0.45} \hlcov{that}{0.00} \hlcov{he}{20.79} \hlcov{plans}{28.92} \hlcov{to}{32.73} \hlcov{aggressively}{25.16} \hlcov{fight}{37.52} \hlcov{corruption}{31.36} \hlcov{that}{22.89} \hlcov{has}{40.00} \hlcov{long}{32.10} \hlcov{plagued}{33.37} \hlcov{nigeria}{34.44} \hlcov{and}{30.59} \hlcov{go}{0.02} \hlcov{after}{0.04} \hlcov{the}{1.03} \hlcov{root}{0.25} \hlcov{of}{0.00} \hlcov{the}{0.10} \hlcov{nation}{0.02} \hlcov{'s}{0.02} \hlcov{unrest}{0.04} \hlcov{.}{0.05} \hlcov{\textit{buhari}}{12.76} \hlcov{said}{6.65} \hlcov{he}{12.70} \hlcov{'ll}{2.57} \hlcov{``}{0.86} \hlcov{rapidly}{0.29} \hlcov{give}{0.85} \hlcov{attention}{0.00} \hlcov{''}{0.00} \hlcov{to}{0.08} \hlcov{curbing}{0.93} \hlcov{violence}{0.85} \hlcov{in}{1.94} \hlcov{the}{0.10} \hlcov{northeast}{0.02} \hlcov{part}{0.02} \hlcov{of}{0.00} \hlcov{nigeria}{1.58} \hlcov{,}{0.85} \hlcov{where}{0.02} \hlcov{the}{1.18} \hlcov{terrorist}{0.97} \hlcov{group}{0.03} \hlcov{boko}{1.40} \hlcov{haram}{0.00} \hlcov{operates}{0.00} \hlcov{.}{0.00} \hlcov{by}{0.40} \hlcov{cooperating}{0.01} \hlcov{with}{0.00} \hlcov{neighboring}{0.05} \hlcov{nations}{0.00} \hlcov{chad}{0.02} \hlcov{,}{0.00} \hlcov{cameroon}{0.28} \hlcov{and}{0.00} \hlcov{niger}{0.00} \hlcov{,}{0.00} \hlcov{he}{5.25} \hlcov{said}{12.55} \hlcov{his}{24.51} \hlcov{administration}{33.19} \hlcov{is}{32.26} \hlcov{confident}{33.16} \hlcov{it}{32.20} \hlcov{will}{32.81} \hlcov{be}{32.56} \hlcov{able}{32.73} \hlcov{to}{32.75} \hlcov{thwart}{32.47} \hlcov{criminals}{32.02} \hlcov{and}{32.03} \hlcov{others}{0.17} \hlcov{contributing}{0.57} \hlcov{to}{0.00} \hlcov{nigeria}{1.45} \hlcov{'s}{0.72} \hlcov{instability}{0.09} \hlcov{.}{5.93} \hlcov{for}{2.13} \hlcov{the}{0.09} \hlcov{first}{0.11} \hlcov{time}{0.00} \hlcov{in}{0.04} \hlcov{nigeria}{1.15} \hlcov{'s}{0.02} \hlcov{history}{0.02} \hlcov{,}{0.22} \hlcov{the}{5.49} \hlcov{opposition}{17.78} \hlcov{defeated}{0.01} \hlcov{the}{0.45} \hlcov{ruling}{0.51} \hlcov{party}{0.04} \hlcov{in}{0.09} \hlcov{democratic}{0.24} \hlcov{elections}{0.02} \hlcov{.}{0.08} \hlcov{\textit{buhari}}{5.74} \hlcov{defeated}{3.55} \hlcov{incumbent}{2.04} \hlcov{goodluck}{0.99} \hlcov{jonathan}{0.28} \hlcov{by}{0.00} \hlcov{about}{0.41} \hlcov{2}{0.05} \hlcov{million}{0.00} \hlcov{votes}{0.00} \hlcov{,}{0.00} \hlcov{according}{0.29} \hlcov{to}{0.01} \hlcov{nigeria}{1.59} \hlcov{'s}{0.00} \hlcov{independent}{0.03} \hlcov{national}{0.02} \hlcov{electoral}{0.01} \hlcov{commission}{0.00} \hlcov{.}{0.03} \hlcov{the}{6.06} \hlcov{win}{20.32} \hlcov{comes}{33.02} \hlcov{after}{32.62} \hlcov{a}{20.84} \hlcov{long}{29.10} \hlcov{history}{33.08} \hlcov{of}{32.51} \hlcov{military}{32.23} \hlcov{rule}{32.95} \hlcov{,}{32.14} \hlcov{coups}{31.18} \hlcov{and}{32.72} \hlcov{botched}{32.77} \hlcov{attempts}{32.89} \hlcov{at}{33.02} \hlcov{democracy}{32.79} \hlcov{in}{26.80} \hlcov{africa}{31.11} \hlcov{'s}{32.57} \hlcov{most}{32.69} \hlcov{populous}{32.76} \hlcov{nation}{32.74} \hlcov{.}{32.51}
\newline \rule{\columnwidth}{0.4pt}
\textbf{Reference Summary:} \newline \textit{muhammadu} \textit{buhari} tells cnn 's christiane amanpour that he will fight corruption in nigeria .\newline nigeria is the most populous country in africa and is grappling with violent boko haram extremists .\newline nigeria is also africa 's biggest economy , but up to 70 \% of nigerians live on less than a dollar a day .
\newline \rule{\columnwidth}{0.4pt}
\textbf{Baseline:} \newline
UNK UNK says his administration is confident it will be able to \error{destabilize nigeria 's economy} . \newline
UNK says his administration is confident it will be able to thwart criminals and other \error{nigerians} . \newline
\error{he says the country has long nigeria and nigeria 's economy .}
\newline \rule{\columnwidth}{0.4pt}
\textbf{Pointer-Generator, No Coverage:} \newline
\textit{muhammadu} \textit{buhari} says he plans to aggressively fight corruption in the northeast part of nigeria . \newline
he says he 'll `` rapidly give attention '' to curbing violence in the northeast part of nigeria . \newline
he says his administration is confident it will be able to thwart criminals .
\newline \rule{\columnwidth}{0.4pt}
\textbf{Pointer-Generator, With Coverage:} \newline  \hlpgen{\textit{muhammadu}}{11.77} \hlpgen{\textit{buhari}}{6.67} \hlpgen{says}{37.60} \hlpgen{he}{8.35} \hlpgen{plans}{17.88} \hlpgen{to}{22.28} \hlpgen{aggressively}{9.39} \hlpgen{fight}{7.52} \hlpgen{corruption}{3.30} \hlpgen{that}{19.02} \hlpgen{has}{9.87} \hlpgen{long}{6.64} \hlpgen{plagued}{3.30} \hlpgen{nigeria}{4.07} \hlpgen{.}{32.51}\newline \hlpgen{he}{13.90} \hlpgen{says}{34.72} \hlpgen{his}{7.75} \hlpgen{administration}{3.11} \hlpgen{is}{9.71} \hlpgen{confident}{3.26} \hlpgen{it}{8.47} \hlpgen{will}{4.98} \hlpgen{be}{3.39} \hlpgen{able}{3.11} \hlpgen{to}{9.68} \hlpgen{thwart}{5.91} \hlpgen{criminals}{3.36} \hlpgen{.}{25.35}\newline \hlpgen{the}{19.76} \hlpgen{win}{6.47} \hlpgen{comes}{21.76} \hlpgen{after}{11.42} \hlpgen{a}{2.89} \hlpgen{long}{1.38} \hlpgen{history}{1.06} \hlpgen{of}{4.70} \hlpgen{military}{1.98} \hlpgen{rule}{1.04} \hlpgen{,}{9.97} \hlpgen{coups}{1.19} \hlpgen{and}{17.53} \hlpgen{botched}{3.51} \hlpgen{attempts}{6.37} \hlpgen{at}{14.83} \hlpgen{democracy}{2.36} \hlpgen{in}{23.41} \hlpgen{africa}{3.47} \hlpgen{'s}{17.72} \hlpgen{most}{2.10} \hlpgen{populous}{1.14} \hlpgen{nation}{1.68} \hlpgen{.}{32.96}\newline
\end{boxedminipage}
\caption{The baseline model incorrectly changes \textit{thwart criminals and others contributing to nigeria's instability} to \textit{destabilize nigeria's economy} -- which has a mostly opposite meaning. It also produces a nonsensical sentence.
Note that our final model produces the novel word \textit{says} to paraphrase \textit{told cnn `s christiane amanpour}.
}
\label{fig_nigeria}
\end{figure*}

\begin{figure*}
\begin{boxedminipage}{\textwidth}
\textbf{Article:}  \hlcov{cairo}{0.07} \hlcov{(}{0.00} \hlcov{cnn}{0.12} \hlcov{)}{0.00} \hlcov{at}{1.45} \hlcov{least}{0.00} \hlcov{12}{0.75} \hlcov{people}{0.09} \hlcov{were}{7.63} \hlcov{killed}{7.67} \hlcov{sunday}{3.82} \hlcov{,}{0.01} \hlcov{and}{0.08} \hlcov{more}{0.74} \hlcov{injured}{0.21} \hlcov{,}{0.10} \hlcov{in}{0.16} \hlcov{separate}{0.11} \hlcov{attacks}{0.36} \hlcov{on}{0.15} \hlcov{a}{0.41} \hlcov{police}{0.62} \hlcov{station}{2.10} \hlcov{,}{9.09} \hlcov{a}{0.19} \hlcov{checkpoint}{0.30} \hlcov{and}{0.06} \hlcov{along}{0.14} \hlcov{a}{0.17} \hlcov{highway}{0.37} \hlcov{in}{0.58} \hlcov{egypt}{0.92} \hlcov{'s}{0.18} \hlcov{northern}{0.13} \hlcov{sinai}{0.06} \hlcov{,}{0.04} \hlcov{authorities}{7.53} \hlcov{said}{0.00} \hlcov{.}{0.76} \hlcov{six}{22.27} \hlcov{people}{30.44} \hlcov{,}{15.07} \hlcov{including}{29.83} \hlcov{one}{31.66} \hlcov{civilian}{30.80} \hlcov{,}{30.86} \hlcov{were}{31.56} \hlcov{killed}{26.72} \hlcov{when}{26.14} \hlcov{a}{30.58} \hlcov{car}{22.37} \hlcov{bomb}{40.00} \hlcov{exploded}{31.02} \hlcov{near}{21.64} \hlcov{the}{29.76} \hlcov{police}{28.96} \hlcov{station}{29.51} \hlcov{in}{25.98} \hlcov{\textit{al-arish}}{0.21} \hlcov{,}{0.01} \hlcov{capital}{0.02} \hlcov{of}{0.00} \hlcov{north}{0.23} \hlcov{sinai}{0.02} \hlcov{,}{0.13} \hlcov{health}{2.39} \hlcov{ministry}{0.05} \hlcov{spokesman}{0.04} \hlcov{\textit{hossam}}{0.04} \hlcov{\textit{abdel-ghafar}}{0.02} \hlcov{told}{0.01} \hlcov{ahram}{0.09} \hlcov{online}{0.07} \hlcov{.}{0.12} \hlcov{he}{2.45} \hlcov{said}{0.23} \hlcov{40}{3.60} \hlcov{people}{0.01} \hlcov{were}{0.76} \hlcov{injured}{1.07} \hlcov{.}{0.17} \hlcov{ansar}{6.47} \hlcov{beit}{21.33} \hlcov{\textit{al-maqdis}}{25.99} \hlcov{,}{23.79} \hlcov{an}{28.92} \hlcov{isis}{34.21} \hlcov{affiliate}{30.95} \hlcov{,}{30.63} \hlcov{claimed}{33.81} \hlcov{responsibility}{30.56} \hlcov{for}{30.64} \hlcov{the}{30.50} \hlcov{attack}{29.70} \hlcov{,}{30.90} \hlcov{which}{0.01} \hlcov{came}{0.05} \hlcov{hours}{0.01} \hlcov{after}{0.00} \hlcov{another}{0.05} \hlcov{operation}{0.01} \hlcov{that}{0.00} \hlcov{the}{0.52} \hlcov{group}{0.14} \hlcov{also}{0.02} \hlcov{claimed}{0.03} \hlcov{.}{0.01} \hlcov{in}{1.85} \hlcov{that}{0.01} \hlcov{earlier}{2.26} \hlcov{attack}{0.14} \hlcov{,}{0.01} \hlcov{a}{1.15} \hlcov{first}{0.22} \hlcov{lieutenant}{0.11} \hlcov{,}{0.00} \hlcov{a}{0.57} \hlcov{sergeant}{0.23} \hlcov{and}{0.00} \hlcov{four}{0.05} \hlcov{\textit{conscripts}}{0.02} \hlcov{were}{0.07} \hlcov{killed}{0.67} \hlcov{when}{0.52} \hlcov{their}{0.43} \hlcov{armored}{0.03} \hlcov{vehicle}{0.02} \hlcov{was}{0.02} \hlcov{attacked}{0.02} \hlcov{on}{0.02} \hlcov{the}{0.01} \hlcov{highway}{0.01} \hlcov{from}{0.02} \hlcov{\textit{al-arish}}{0.03} \hlcov{to}{0.00} \hlcov{sheikh}{0.07} \hlcov{\textit{zuweid}}{0.00} \hlcov{in}{0.01} \hlcov{northern}{0.01} \hlcov{sinai}{0.01} \hlcov{,}{0.00} \hlcov{the}{0.21} \hlcov{military}{0.22} \hlcov{said}{0.04} \hlcov{.}{0.01} \hlcov{two}{3.04} \hlcov{other}{0.03} \hlcov{soldiers}{0.41} \hlcov{were}{0.34} \hlcov{injured}{0.34} \hlcov{and}{0.02} \hlcov{taken}{0.02} \hlcov{to}{0.00} \hlcov{a}{0.00} \hlcov{military}{0.02} \hlcov{hospital}{0.00} \hlcov{.}{2.41} \hlcov{ansar}{9.54} \hlcov{beit}{9.43} \hlcov{\textit{al-maqdis}}{5.11} \hlcov{has}{2.89} \hlcov{claimed}{1.07} \hlcov{many}{0.30} \hlcov{attacks}{0.12} \hlcov{against}{0.01} \hlcov{the}{0.03} \hlcov{army}{0.08} \hlcov{and}{0.01} \hlcov{police}{0.09} \hlcov{in}{0.08} \hlcov{sinai}{0.01} \hlcov{.}{0.01} \hlcov{a}{2.97} \hlcov{third}{0.10} \hlcov{attack}{0.06} \hlcov{sunday}{0.35} \hlcov{on}{0.01} \hlcov{a}{0.02} \hlcov{checkpoint}{0.00} \hlcov{in}{0.01} \hlcov{rafah}{0.02} \hlcov{left}{0.01} \hlcov{three}{0.13} \hlcov{security}{0.03} \hlcov{personnel}{0.00} \hlcov{injured}{0.02} \hlcov{,}{0.00} \hlcov{after}{0.01} \hlcov{unknown}{0.11} \hlcov{assailants}{0.07} \hlcov{opened}{0.00} \hlcov{fire}{0.00} \hlcov{at}{0.01} \hlcov{them}{0.01} \hlcov{,}{0.03} \hlcov{according}{0.00} \hlcov{to}{0.00} \hlcov{state}{0.21} \hlcov{media}{0.00} \hlcov{.}{0.00} \hlcov{the}{5.60} \hlcov{attacks}{2.17} \hlcov{come}{0.00} \hlcov{as}{0.00} \hlcov{the}{0.52} \hlcov{military}{0.15} \hlcov{announced}{0.05} \hlcov{a}{0.00} \hlcov{reshuffle}{0.00} \hlcov{of}{0.00} \hlcov{several}{0.01} \hlcov{senior}{0.01} \hlcov{military}{0.00} \hlcov{positions}{0.00} \hlcov{,}{0.00} \hlcov{state}{1.17} \hlcov{media}{0.01} \hlcov{reported}{0.00} \hlcov{.}{0.01} \hlcov{among}{6.73} \hlcov{those}{0.04} \hlcov{being}{0.00} \hlcov{replaced}{0.01} \hlcov{are}{0.00} \hlcov{the}{0.00} \hlcov{generals}{0.01} \hlcov{in}{0.00} \hlcov{charge}{0.00} \hlcov{of}{0.00} \hlcov{military}{0.12} \hlcov{intelligence}{0.00} \hlcov{and}{0.00} \hlcov{egypt}{0.30} \hlcov{'s}{1.34} \hlcov{second}{0.51} \hlcov{field}{0.00} \hlcov{army}{0.03} \hlcov{,}{0.60} \hlcov{which}{0.01} \hlcov{is}{0.05} \hlcov{spearheading}{0.02} \hlcov{the}{0.00} \hlcov{battle}{0.03} \hlcov{against}{0.01} \hlcov{the}{0.03} \hlcov{insurgents}{0.03} \hlcov{in}{0.00} \hlcov{the}{0.00} \hlcov{northern}{0.01} \hlcov{sinai}{0.00} \hlcov{.}{0.07} \hlcov{egypt}{10.20} \hlcov{'s}{29.24} \hlcov{army}{30.40} \hlcov{has}{30.04} \hlcov{been}{29.23} \hlcov{fighting}{30.87} \hlcov{a}{29.71} \hlcov{decade-long}{26.59} \hlcov{militant}{32.90} \hlcov{islamist}{32.88} \hlcov{insurgency}{31.69} \hlcov{,}{30.55} \hlcov{which}{0.14} \hlcov{has}{0.02} \hlcov{spiked}{1.50} \hlcov{since}{0.07} \hlcov{the}{0.00} \hlcov{ouster}{0.00} \hlcov{of}{0.00} \hlcov{muslim}{0.09} \hlcov{brotherhood}{0.00} \hlcov{president}{0.02} \hlcov{mohamed}{0.00} \hlcov{morsy}{0.00} \hlcov{in}{0.02} \hlcov{the}{0.00} \hlcov{summer}{0.00} \hlcov{of}{0.00} \hlcov{2013}{0.00} \hlcov{.}{0.05} \hlcov{hundreds}{3.39} \hlcov{of}{0.00} \hlcov{police}{0.05} \hlcov{and}{0.00} \hlcov{soldiers}{0.02} \hlcov{,}{0.08} \hlcov{as}{0.00} \hlcov{well}{0.00} \hlcov{as}{0.00} \hlcov{civilians}{0.09} \hlcov{,}{0.05} \hlcov{have}{0.01} \hlcov{been}{0.00} \hlcov{killed}{0.07} \hlcov{in}{0.03} \hlcov{militant}{0.04} \hlcov{attacks}{0.06} \hlcov{in}{0.00} \hlcov{the}{0.00} \hlcov{past}{0.00} \hlcov{months}{0.00} \hlcov{.}{0.00} \hlcov{ian}{13.93} \hlcov{lee}{0.00} \hlcov{reported}{0.00} \hlcov{from}{0.00} \hlcov{cairo}{0.01} \hlcov{.}{0.00} \hlcov{anas}{1.54} \hlcov{hamdan}{0.00} \hlcov{reported}{0.01} \hlcov{from}{0.00} \hlcov{atlanta}{0.02} \hlcov{.}{0.01}                                                                                                                                                                                                  \newline \rule{\columnwidth}{0.4pt}                                                                                                                                                                                              \textbf{Reference Summary:} \newline six people , including one civilian , are killed when a car bomb explodes near a police station .\newline six others are killed when their armored vehicle is attacked on a highway in northern sinai .\newline ansar beit \textit{al-maqdis} , an isis affiliate , claims responsibility .
\newline \rule{\columnwidth}{0.4pt}                                                                                                                                                                                              \textbf{Baseline:} \newline
he says the attacks come after another operation that killed 12 . \newline
the attacks come as military announced a \error{u.n. peacekeeping force} in northern sinai .
 \newline \rule{\columnwidth}{0.4pt}
\textbf{Pointer-Generator, No Coverage:} \newline
ansar beit \textit{al-maqdis} , an isis affiliate , claimed responsibility for the attack . \newline
ansar beit \textit{al-maqdis} , an isis affiliate , claimed responsibility for the attack . \newline
the attacks come as the military announced a reshuffle of several senior military positions .
\newline \rule{\columnwidth}{0.4pt}
\textbf{Pointer-Generator, With Coverage:} \newline  \hlpgen{six}{12.28} \hlpgen{people}{6.00} \hlpgen{,}{27.43} \hlpgen{including}{14.46} \hlpgen{one}{3.29} \hlpgen{civilian}{2.05} \hlpgen{,}{32.11} \hlpgen{were}{17.88} \hlpgen{killed}{9.73} \hlpgen{when}{27.53} \hlpgen{a}{7.57} \hlpgen{car}{0.74} \hlpgen{bomb}{2.74} \hlpgen{explodes}{26.54} \hlpgen{near}{26.47} \hlpgen{the}{3.28} \hlpgen{police}{1.60} \hlpgen{station}{3.77} \hlpgen{.}{29.54}\newline \hlpgen{ansar}{8.00} \hlpgen{beit}{0.50} \hlpgen{\textit{al-maqdis}}{0.67} \hlpgen{,}{15.55} \hlpgen{an}{7.02} \hlpgen{isis}{0.30} \hlpgen{affiliate}{0.25} \hlpgen{,}{22.88} \hlpgen{claimed}{7.66} \hlpgen{responsibility}{2.11} \hlpgen{for}{18.22} \hlpgen{the}{1.11} \hlpgen{attack}{1.98} \hlpgen{.}{35.12}\newline \hlpgen{egypt}{21.26} \hlpgen{'s}{20.48} \hlpgen{army}{4.13} \hlpgen{has}{18.17} \hlpgen{been}{2.88} \hlpgen{fighting}{2.69} \hlpgen{a}{2.93} \hlpgen{decade-long}{1.22} \hlpgen{militant}{1.50} \hlpgen{islamist}{0.94} \hlpgen{insurgency}{3.65} \hlpgen{.}{33.79}\newline
\end{boxedminipage}
\caption{The baseline model fabricates a completely false detail about \textit{a u.n. peacekeeping force} that is not mentioned in the article.
This is most likely inspired by a connection between U.N. peacekeeping forces and \textit{northern sinai} in the training data.
The pointer-generator model is more accurate, correctly reporting the \textit{reshuffle of several senior military positions}.
}
\label{fig_carbomb}
\end{figure*}

\end{document}